\documentclass[journal]{IEEEtran}
                             
\usepackage{cite}
\usepackage{algorithm}  
\usepackage{algpseudocode}
\usepackage{multirow}
\usepackage{amsmath}
\usepackage{amssymb}
\usepackage{hyperref}

\usepackage{epsfig} 
\usepackage{amsmath} 

\usepackage{graphicx}
\usepackage{caption}
\usepackage{subcaption}
\usepackage{amsthm}

\usepackage[flushleft]{threeparttable}

\usepackage{tabularx} 

\usepackage{booktabs}
\usepackage{xcolor}
\usepackage{color}
\usepackage{enumitem}
\usepackage{breakurl}

\title{A Conflicts-free, Speed-lossless KAN-based Reinforcement Learning Decision System for Interactive Driving in Roundabouts}

\author{Zhihao Lin$^{1, \dagger}$,  Zhen Tian$^{1, \dagger}$, Jianglin Lan$^{1}$, Qi Zhang$^{2}$, Ziyang Ye$^{3}$, Hanyang Zhuang$^{4}$,~\IEEEmembership{Member,~IEEE},  and Xianxian Zhao$^{5,\ast}$
\thanks{*This work was supported in part by the China Scholarship Council Ph.D. Scholarship for 2023-2027 (No.202206170011), in part by the Leverhulme Trust Early Career Fellowship (ECF-2021-517), in part by the UK Royal Society International Exchanges Cost Share Programme (IEC$\backslash$NSFC$\backslash$223228), and in part by the SEAI (Sustainable Energy Authority of Ireland) under RD$\&$D Award 22/RDD/776. }
\thanks{$^{1}$Zhihao Lin, Zhen Tian and Jianglin Lan are with James Watt School of Engineering, University of Glasgow, Glasgow G12 8QQ, United Kingdom}%
\thanks{$^{2}$Qi Zhang is with Faculty of Science, University of Amsterdam, Science Park 904, 1098 XH Amsterdam, Netherlands}%
\thanks{$^{3}$Ziyang Ye is with School of Computer and Mathematical Sciences, The University of Adelaide, South Australia 5005, Australia}
\thanks{$^{4}$Hanyang Zhuang is with University of Michigan-Shanghai Jiao Tong University Joint Institute, Shanghai Jiao Tong University,
Shanghai, 200240, China}
\thanks{$^5$Xianxian Zhao is with the School of Electrical and Electronic Engineering, University College Dublin, Belfield, D04 V1W8 Dublin, Ireland}
\thanks{$^{\ast}$Corresponding author. Xianxian Zhao(e-mail: xianxian.zhao@ucd.ie)}%
\thanks{$\dagger$ Equal contribution}
}

\begin{document}
\maketitle

\begin{abstract}
Safety and efficiency are crucial for autonomous driving in roundabouts, especially mixed traffic with both autonomous vehicles (AVs) and human-driven vehicles. This paper presents a learning-based algorithm that promotes safe and efficient driving across varying roundabout traffic conditions. 
A deep Q-learning network is used to learn optimal strategies in complex multi-vehicle roundabout scenarios, while a Kolmogorov-Arnold Network (KAN) improves the AVs' environmental understanding. To further enhance safety, an action inspector filters unsafe actions, and a route planner optimizes driving efficiency. Moreover, model predictive control ensures stability and precision in execution. Experimental results demonstrate that the proposed system consistently outperforms state-of-the-art methods, achieving fewer collisions, reduced travel time, and stable training with smooth reward convergence.
\end{abstract}



\section{Introduction}
\IEEEPARstart{R}{oundabout} designs vary by city scale~\cite{Ladosz2024}, but typically feature a central island that vehicles must circulate around—clockwise or counterclockwise—facilitating smoother traffic flow and reducing interaction complexity~\cite{HOU2025104916}.
As urban roadways evolve, roundabouts have improved traffic distribution and increased road capacity~\cite{MA2024103014}. While they generally present fewer conflicts than other intersections~\cite{Wang2023}, safety concerns intensify in high-traffic conditions due to a greater crash risk~\cite{WU2024103069}, especially in mixed traffic with both human-driven vehicles (HDVs) and autonomous vehicles (AVs). 

Understanding human driving behavior in roundabouts, especially during entering, circulating, and exiting, is a key research focus. Merging and exiting require AVs to interact with surrounding HDVs, interpreting their intentions to make optimized decisions. Lane changes demand careful monitoring of HDVs and precise control to follow planned trajectories. Therefore, AVs navigating roundabouts must select lanes appropriately, monitor their environment, avoid collisions, and maintain precise control.

Research on autonomous driving in roundabouts has advanced from basic navigation to complex mixed-traffic scenarios. AVs reduce safety incidents caused by human errors like fatigue and distraction~\cite{BADUE2021113816} and can make faster, optimal decisions~\cite{MAK2023104376}. They also enhance roundabout capacity~\cite{Xing2024}. With AVs expected to exceed 50 million on the road by 2030~\cite{PatentPC2023}, modern roundabouts are increasingly designed to support safe AV-HDV interactions~\cite{YU2025104944,Xi2024}. Control strategies for connected autonomous vehicles prioritize safety and efficiency~\cite{ma2023distributed}, while current roundabout designs improve traffic flow and safety~\cite{Chen2024}.

Control methods for autonomous driving in roundabouts have gained significant attention, with Model Predictive Control (MPC) and game theory being prominent model-based approaches. Game theory models decision-making by balancing safety, efficiency, and comfort~\cite{hang2020integrated}, but often relies on simplified environments and struggles with real-world complexity~\cite{9831031}. Other model-based frameworks are limited by simplistic roundabout designs, few vehicles, and focus on abnormal cases~\cite{9907889}. While MPC effectively handles vehicle dynamics and safety constraints, current approaches still face challenges in complex real-world roundabouts~\cite{Mao2023,9177294}.

Learning-based methods, including machine learning and deep reinforcement learning (DRL), show strong potential for complex roundabout driving. Machine learning has been applied to AV-HDV interactions~\cite{tian2020game}, but often requires extensive labeled data and struggles with generalization. DRL enables exploration of strategies in complex environments~\cite{Shi2024} and balances safety and efficiency in dense traffic~\cite{Cai2024}. Popular DRL algorithms include Deep Deterministic Policy Gradient (DDPG)\cite{Lei2024}, Proximal Policy Optimization (PPO)\cite{9693175}, and deep Q-learning (DQN)\cite{Li2024}. DDPG suits continuous actions but is less effective for discrete decisions like roundabout driving\cite{basile2022ddpg}. PPO achieves safe, efficient strategies in dense traffic~\cite{Chen2024} but its on-policy learning limits use of historical data, which is vital in complex environments like roundabouts~\cite{Liu2023}
DQN has been effectively applied to traffic simulations, excelling in tasks like intersection management relevant to roundabout navigation~\cite{Liu2024}. Its discrete action framework suits lane selection without action discretization, and experience replay enhances learning efficiency using past data. Additionally, DQN is computationally efficient~\cite{cai2022dq}. The recently proposed Kolmogorov-Arnold Network (KAN) outperforms traditional multi-layer perceptrons by replacing linear layers with adaptive B-spline functions, enabling flexible feature extraction and improved generalization across diverse environments~\cite{liu2024kan,Kundu2024}.

To solve the complex driving in roundabouts, this paper proposes to integrate KAN with DQN (K-DQN) to enhance the decision-making and learning capabilities of AVs in complex roundabout scenarios~\cite{MALLICK2024111803}. The K-DQN leverages the advantages of both DQN and KAN, enabling AVs to learn robust and efficient driving strategies through interaction with the environment. For conflict-free driving, we introduce an action inspector applied to time to collision (TTC)~\cite{Bie2024} to assess the relative collision risks between the AV and other HDVs. By replacing dangerous actions that may cause collisions with safe actions, our proposed method can decrease the ego vehicle collision rates with neighboring vehicles (NVs) during training. For proper lane selection, we introduce a route planner that considers the number of HDVs and the available free-driving space in each lane. For precise control of planned trajectories, we implement MPC to allow the AV~\cite{KEMPF2024111782} to navigate with precision and robustness.

The main contributions of this paper are as follows:
\begin{itemize}
    \item We propose a novel K-DQN to enhance AV decision-making in complex roundabouts.  Compared to the traditional neural networks, the unique spline-based activation functions of KAN enable more precise environmental learning and decision-making, resulting in better training convergence, lower collision rates, and higher average speeds.

    \item Unlike prior methods that treat safety and efficiency separately, we introduce an integrated approach combining an action inspector and route planner. By merging TTC-based safety checks with density-aware lane selection, our method significantly reduces collisions while improving driving efficiency across diverse traffic conditions, compared to benchmarks.

    \item We enhance traditional DRL by integrating MPC with K-DQN, translating planned actions into safe, smooth controls.  Our integrated solution adeptly manages diverse roundabout traffic flows, showing improved speed stability and efficiency over current benchmarks.

    \item We present mathematical analysis and experimental demonstrations to substantiate the superior performance of our K-DQN over traditional DQN methods. Extensive simulations confirm robustness and efficiency of our approach and its advantages over benchmarks.
\end{itemize}

The rest of this paper is organized as follows: Section~\ref{sec2} presents the problem statement and system structure; Section~\ref{sec3} describes the enhanced K-DQN; Section~\ref{sec4} introduces the action inspector and route planner; Section~\ref{sec5} presents the MPC design; Section~\ref{sec6} provides the simulation results with analysis; Section~\ref{sec7} draws the conclusion.



\begin{figure}[t]
    \centering
    \includegraphics[width=0.7\linewidth]{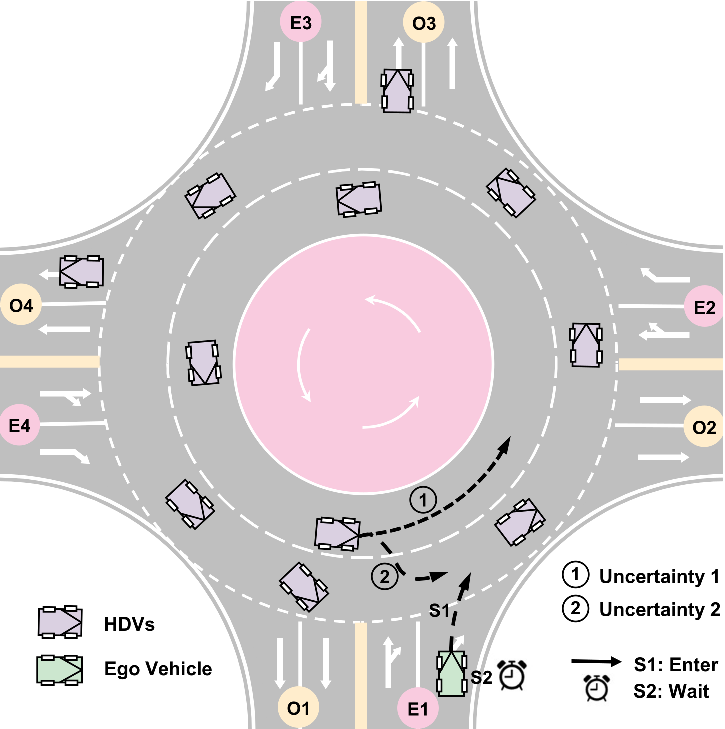}
    \caption{A four-entrance, four-outlet, two-lane roundabout with the first collision scenario involving AV uncertainty.}
    \label{fig1}
\end{figure}

\begin{figure}[t] 
    \centering
    \begin{subfigure}[b]{0.22\textwidth}
        \includegraphics[width=\linewidth]{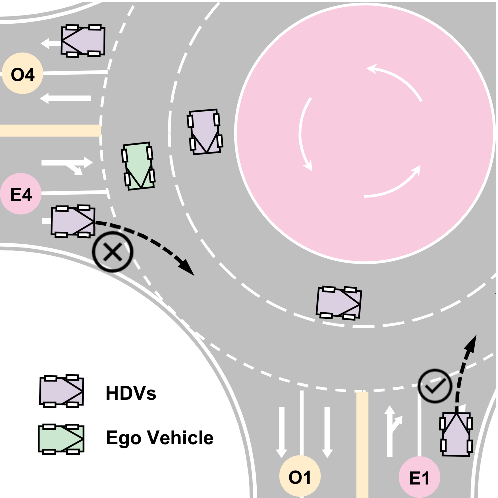}
        \caption{}
        \label{fig2a}
    \end{subfigure}%
    \hspace{5pt} 
    \begin{subfigure}[b]{0.22\textwidth}
        \includegraphics[width=\linewidth]{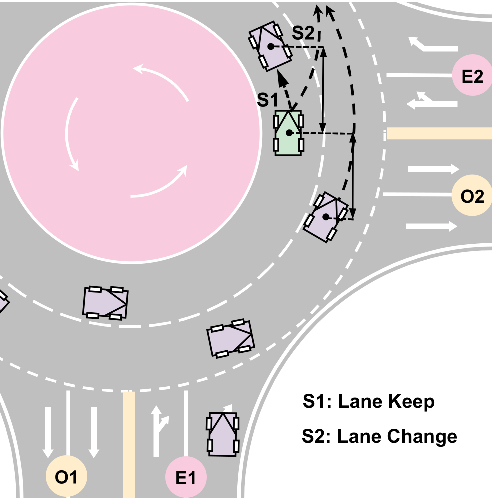}
        \caption{}
        \label{fig2b}
    \end{subfigure}%
    \caption{Roundabout potential collision scenarios (a) and (b).}
    \label{fig2}
\end{figure}

\begin{figure*}[t]
    \centering
    \includegraphics[width=0.7\linewidth]{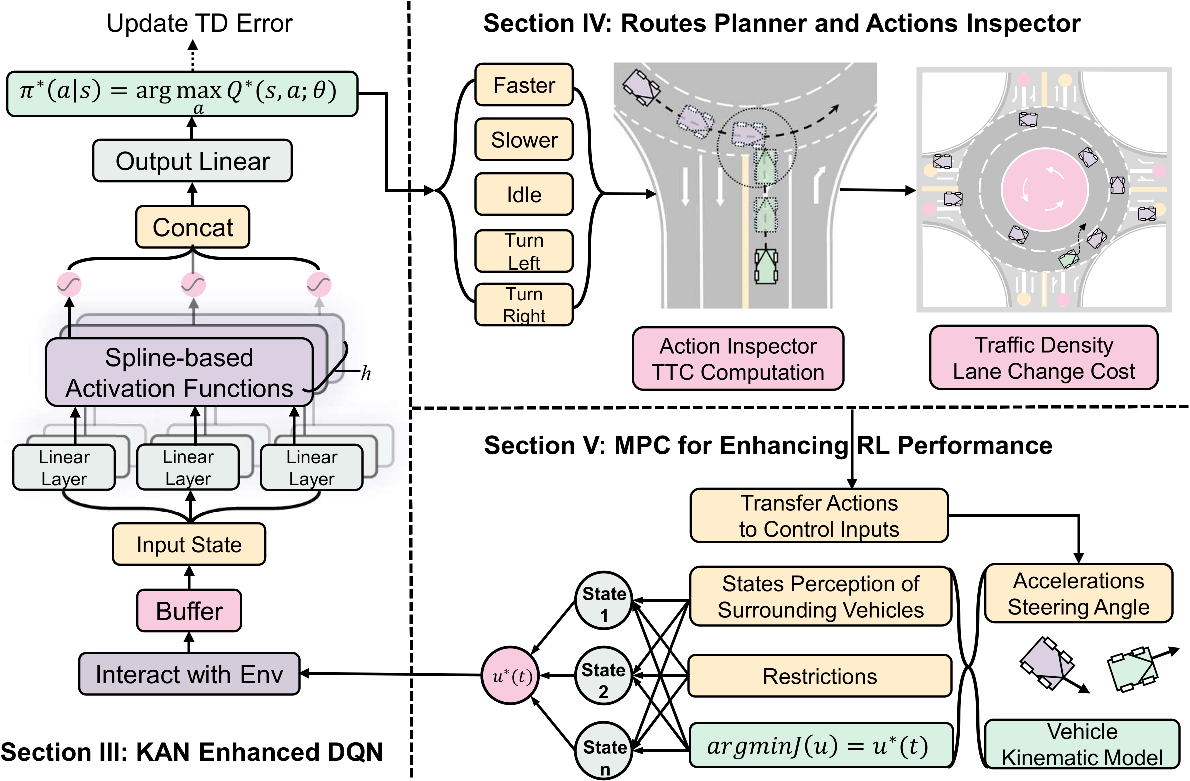}
    \caption{The KAN-based, conflict-avoidance, and proper-lane-detection DRL system.}
    \label{fig3}
\end{figure*}

\section{Problem Statement and System Structure}
\label{sec2}
The previous section highlighted the focus on AV-HDV interaction in roundabouts. However, integrating decision-making, path planning, and control remains challenging due to the complexity of roundabout scenarios. Unlike straight or other curvy roads, roundabouts present unique challenges in making safe and efficient decisions due to their complex network of entrances and outlets. HDVs can be randomly and densely distributed along both the inner and outer boundaries of roundabouts, frequently resulting in unexpected outcomes such as conflicts and inefficient driving. As shown in Fig.~\ref{fig1}, the roundabout has four ports, each split into an entrance (right) and an outlet (left). HDVs' unpredictable maneuvers and unknown destinations pose challenges for AVs to ensure both safety and efficiency, defined here as minimizing travel time to the outlet. This study considers a signal-free, double-lane roundabout with both AVs and two types of HDVs: those already inside and those merging in.

Three potential collision situations have been identified in Fig.~\ref{fig1}, Fig.~\ref{fig2}(a), and Fig.~\ref{fig2}(b), respectively.
In Fig.~\ref{fig1}, an AV at an entrance must decide whether to wait for an approaching HDV in the outer lane—potentially increasing delay—or merge immediately, risking a collision due to limited distance and uncertain HDV speed. Fig.~\ref{fig2}(a) presents the reverse case: the AV is approaching an entrance while an HDV attempts to merge, creating a conflict if both proceed simultaneously. In Fig.~\ref{fig2}(b), the AV intends to exit from the inner lane but encounters an HDV in the adjacent outer lane. It must choose between overtaking with risk, following to reduce conflict, or delaying the lane change. These scenarios highlight the complexity and uncertainty AVs face in interactive decision-making with HDVs in roundabouts.

This paper focuses on ensuring the safety and efficiency of AVs navigating roundabouts under varying HDV densities and traffic flows. The proposed system, illustrated in Fig.~\ref{fig3}, addresses this by integrating adaptive decision-making, safety assurance, and robust control. The system comprises four components: environment, decision network, safety-efficiency mechanism, and robust control. The environment updates AV states and computes rewards based on control commands. The decision network selects actions that balance safety and efficiency. The safety-efficiency mechanism includes a route planner, which assists in lane selection during merging, and an action inspector that filters unsafe actions, especially in interactions with HDVs. For control execution, MPC ensures that the chosen actions are translated into smooth and reliable commands. In emergency scenarios, the action inspector identifies approaching emergency vehicles and triggers appropriate responses, such as yielding or changing lanes. The route planner concurrently adjusts the AV’s trajectory to minimize interference, ensuring compliance with safety norms.

\begin{table}[t]
\centering
\setlength{\tabcolsep}{5pt}
\captionsetup{
    labelfont={sc}, 
    textfont={sc}, 
    labelsep=colon, 
    skip=1em,     
    singlelinecheck=false,
    justification=centering,
    format=plain
}
\caption{Variables and description}
\label{tab:variables}
\begin{tabular}{c|l}
\hline\hline
\textbf{Variable} & \textbf{Description} \\
\hline
$s_t, a_t, r_t$ & State, action, and reward at time step $t$ \\
$Q(s,a;\theta)$ & Approximate action-value function with parameters $\theta$ \\
$Q^*(s,a)$ & Optimal action-value function \\
$\theta$, $\theta'$ & Parameters of the current Q-network and target Q-network \\
$\alpha_i, \beta_i$ & Learnable coefficients in the KAN activation function \\
$\lambda_1, \lambda_2$ & Regularization coefficients in the KAN architecture \\
$\mathcal{L}(\theta)$ & Loss function for training the Q-network \\
$\gamma$ & Discount factor \\
$\mathbb{E}$ & Expectation operator \\
$f(x)$ & Output from the KAN layer \\
$W$, $b$ & Weight and bias of the output layer \\
$j$ & Index of the output layer neuron \\
$n$ & Total number of output layer neurons \\
$\Phi_{l,i,j}$ & Spline functions in the approximation theory \\
$\Phi^G_{l,i,j}$ & $k$-th order B-spline functions in the approximation theory \\
$C$, $G$ & Constant and Grid size in the approximation theory \\
$L_b$, $L_{\text{spline}}$ & Lipschitz constants for activation, and spline functions \\
$L_{Q^*}$ & Lipschitz constant for the optimal action-value function \\
$\varepsilon$ & Approximation error \\
$\alpha_{\max}, \beta_{\max}$ & Maximum values of $\alpha_i$ and $\beta_i$ \\
\hline\hline
\end{tabular}
\end{table}

\section{KAN-enhanced DQN Method}
\label{sec3}
The K-DQN network consists of a replay memory, a KAN-Q-network, and a target Q-network. The KAN-Q-network processes environmental data to compute Q-values for safe, efficient decision-making, leveraging robust and precise learning. The target Q-network shares the same structure but updates less frequently to reduce learning instability. Its parameters are periodically synchronized with the KAN-Q-network. Key variables used in the K-DQN's mathematical derivations are summarized in Table~\ref{tab:variables}.

\subsection{Basic DQN}
DQN integrates deep learning with Q-learning to handle high-dimensional state spaces. It uses a neural network to approximate the optimal action-value function $Q^*(s,a)$—the maximum expected return for taking action $a$ in state $s$. DQN relies on experience replay and target networks. Experience replay stores transitions $(s_t, a_t, r_t, s_{t+1})$ for repeated learning, enhancing stability. The target network, updated periodically from the main Q-network, computes target Q-values to reduce correlations and stabilize training.

Let $Q(s_t, a_t)$ denote the Q-value for taking action $a_t$ in state $s_t$ at time step $t$. In Q-learning, the update rule for the Q-value is given by
\begin{equation}
\!\!Q(s_t, \!a_t) \!\leftarrow\! Q(s_t, \!a_t) + \alpha (r_t + \gamma \max_a Q(s_{t+1}, \!a) - Q(s_t,\! a_t)) \!\!\!
\end{equation}
where $\alpha$ is the learning rate, $r_t$ is the immediate reward, $\gamma$ is the discount factor, and $\max_{a} Q(s_{t+1}, a)$ is the maximum Q-value over actions in the next state $s_{t+1}$.

Let $y$ be the ideal (target) Q-value of the current action calculated from the Bellman equation at time step $t$.
During the training process, $y$ is computed by
\begin{equation}\label{eq7}
y = r_t + \gamma \max_{a'} Q(s_{t+1}, a'; \theta^{\prime})
\end{equation}
where $\max_{a'} Q(s_{t+1}, a'; \theta^{\prime})$ is the maximum Q-value over actions $a'$ in the next state $s_{t+1}$, estimated with parameters $\theta^{\prime}$. Note that the subscript $t$ denotes values at time $t$, while unsubscripted variables are general.

The loss function is defined on the difference between $Q(s_t, a_t; \theta)$ and $y$ as follows:
\begin{equation}
\mathcal{L}(\theta) = \mathbb{E}\left[(y - Q(s_t,a_t;\theta))^2\right]
\label{eq3-1}
\end{equation}
where $\theta$ is the target network parameter, and $\mathbb{E}$ is the expectation over all state-action pairs $(s_t, a_t)$ during training.

In the DQN framework, the goal is to minimize the loss function $\mathcal{L}(\theta)$. After computing the loss, the Q-network’s weights are updated by descending the gradient to reduce $\mathcal{L}$. The gradient $\frac{\partial \mathcal{L}}{\partial \theta}$ used for this update is given by 
 \begin{equation}\label{eq4}
 \frac{\partial \mathcal{L}}{\partial \theta}=\mathbb{E} \left[2 (Q(s_t, a_t; \theta) - y) \frac{\partial Q}{\partial \theta} \right].
\end{equation}


Gradients are essential for updating the DQN parameters, as they measure how the loss function changes with respect to the Q-network’s parameters and indicate the direction of steepest descent. By computing these gradients, we identify how to adjust the parameters to effectively minimize the loss. Specifically, we update the parameters $\theta$ using gradient descent by moving in the negative gradient direction:
\begin{equation}
\theta \leftarrow \theta - \alpha \frac{\partial \mathcal{L}}{\partial \theta}
\end{equation}
where $\alpha$ is the learning rate controlling the update step size. This process repeats until the loss converges, aligning the Q-network’s predictions with the targets.

Basic DQN faces challenges in complex environments like roundabouts: training instability from correlated samples and moving targets, the exploration-exploitation trade-off, sensitivity to hyperparameters, and overestimation bias causing suboptimal policies.
To balance exploration and exploitation, we use a dynamic $\varepsilon$-greedy strategy, gradually reducing $\varepsilon$ from 0.9 to 0.1 for broad exploration and stable refinement. KAN’s adaptive spline activations enhance feature representation and reduce hyperparameter sensitivity, improving robustness. Overestimation bias is mitigated by KAN’s accurate Q-value approximation and the action inspector’s real-time safety checks, preventing unsafe actions from skewing policy updates.

\subsection{Structure of KAN}
\label{sec3b}
KAN’s core uses spline-based activation functions of the form:
\begin{equation}
f(x_i; \theta_i, \beta_i, \alpha_i) = \alpha_i \cdot \text{spline}(x_i; \theta_i) + \beta_i \cdot b(x_i)
\label{eqsk}
\end{equation}
where $x_i$ is the input to the $i$-th neuron, $\text{spline}(x_i; \theta_i)$ represents the spline function parameterized by coefficient $\theta_i$, $b(x_i) = \text{SiLU}(x_i) = x_i/(1 + e^{-x_i})$ is an activation function, and $\alpha_i$ and $\beta_i$ are learnable coefficients. Spline functions are piecewise polynomials that can approximate any continuous function. By tuning their parameters, KAN can model complex nonlinear functions.

The coefficient $\theta_i$ is updated via gradient descent on the loss $\mathcal{L}(\theta)$ in~\eqref{eq3-1}, following the update rule:
\begin{equation}
\theta_i^{(t+1)} = \theta_i^{(t)} - \eta \frac{\partial \mathcal{L}}{\partial \theta_i}
\end{equation}
with the learning rate $\eta$. 

KAN uses regularization to reduce overfitting by adding to $\mathcal{L}(\theta)$ the term:
\begin{equation}
\mathcal{R}(\theta) = \lambda_1 \sum_i |\theta_i| + \lambda_2 \sum_i \sum_{j \neq i} |\theta_i - \theta_j|
\end{equation}
where $\lambda_1$ and $\lambda_2$ are regularization coefficients. The $L_1$ term $\lambda_1 \sum_i |\theta_i|$ promotes sparsity, while $\lambda_2 \sum_i \sum_{j \neq i} |\theta_i - \theta_j|$ enforces smoothness across neurons, enhancing stability. Overall, adding $\mathcal{R}(\theta)$ controls model complexity and improves generalization. 

KAN also uses parameter sharing among neurons, defined as:
\begin{equation}
\theta_{\text{shared}} = \frac{1}{N_{\text{group}}} \sum_{i \in \text{group}} \theta_i
\end{equation}
where $\text{group}$ indexes neurons sharing parameters, and $N_{\text{group}}$ is the group size. Shared parameters $\theta_{\text{shared}}$ average neuron parameters, reducing model complexity, improving efficiency, and enhancing generalization.

These elements of the KAN architecture collectively enhance the flexibility and efficiency of the learning process, whilst ensuring robustness against overfitting and maintaining high performance across reinforcement learning tasks.

\subsection{KAN Enhanced DQN}
Integrating KAN into DQN (K-DQN) enhances Q-function approximation, boosting learning robustness and policy performance in complex DRL tasks. To justify pairing KAN with DQN over other RL methods, it’s crucial to analyze their differences in handling environments and KAN’s activation function traits.

To model roundabout driving as a Markov Decision Process, we define these key components:

\textbf{State Space $\mathcal{S}$:} It consists of the ego vehicle's position, velocity, and heading, as well as the relative positions, velocities, and headings of the surrounding vehicles within a certain range. The state at time $t$ is represented as
\begin{equation}
s_t = [p_\textrm{EV}(t), v_\textrm{EV}(t), h_\textrm{EV}(t), p_\textrm{NV}^i(t), v_\textrm{NV}^i(t), h_\textrm{NV}^i(t)]^\top
\end{equation}
where $p_\textrm{EV}(t)$, $v_\textrm{EV}(t)$, and $h_\textrm{EV}(t)$ denote the position, velocity, and heading of the ego vehicle, while $p_\textrm{NV}^i(t)$, $v_\textrm{NV}^i(t)$, and $h_\textrm{NV}^i(t)$ are the position, velocity, and heading of the $i$-th neighboring vehicle.

\textbf{Action Space $\mathcal{A}$:} It is discrete and consists of five high-level actions: faster, slower, idle, turn right, and turn left.

\textbf{Reward Function $r$:} It encourages the ego vehicle to drive safely and efficiently through the roundabout, designed as:
\begin{equation}
r(s_t, a_t) = w_1 r_\textrm{c} + w_2 r_\textrm{s} + w_3 r_\textrm{l\_c} + w_4 r_\textrm{h} + w_5 r_\textrm{a}
\end{equation}
where $r_\textrm{c}$ assigns a large penalty (-100) for collisions, $r_\textrm{s}$ provides continuous feedback proportional to the vehicle's speed ($v/v_{\text{max}}$), $r_\textrm{l\_c}$ adds a small negative value (-10) for each lane change to prevent unnecessary maneuvers, $r_\textrm{h}$ encourages maintaining safe distances by scaling with inverse headway time, and $r_\textrm{a}$ gives a positive reward (+200) for reaching the target exit. The weights are empirically set as $w_1=1.0$, $w_2=0.3$, $w_3=0.2$, $w_4=0.3$, and $w_5=0.2$ to balance safety with efficiency.

By using \eqref{eqsk}, the goal is to directly approximate the optimal action-value function 
\begin{align}\label{eq12}
&Q(s,a; \theta) = \sum_{j} W_j^T f(x) + b \notag\\
&=\!\!\sum_{j\in [1-n]} \sum_{i=1}^m \left(\alpha_i \cdot \text{spline}(x_i; \theta_i) + \beta_i \!\cdot\! b(x_i)\right) + b \notag\\
&=\!\!\sum_{j\in [1-n]} \sum_{i=1}^{m} (\alpha_i \cdot \text{spline}((s,a)_i; \theta_i) \!+ \beta_i \!\cdot\! b((s,a)_i)) + b \!\!
\end{align}
where $W$ and $b$ are the weight and bias of the network, $f(x)$ is the output from the KAN layer, $j$ is the index of the output layer neuron, and $n$ is the total number of output layer neurons.

\textit{Theorem 1}: (Approximation theory~\cite{liu2024kan})
Suppose that a function $f(x)$ admits a representation
$
f = (\Phi_{L-1} \circ \Phi_{L-2} \circ \cdots \circ \Phi_1 \circ \Phi_0)x
$,
where each part $\Phi_l$ is $(k + 1)$-times continuously differentiable. Then there exist $k$-th order B-spline functions $\Phi^G_{l}$ such that for any $0 \leq m \leq k$, 
\begin{equation}
\|f - (\Phi^G_{L-1} \circ \Phi^G_{L-2} \circ \cdots \circ \Phi^G_1 \circ \Phi^G_0)x\|_{C^m} \leq CG^{-k-1+m}
\end{equation}
where $C$ is a constant and $G$ is the grid size. The magnitude of derivatives up to order $m$ is measured by the $C^m\text{-norm}$ as 
\begin{equation}
\|g\|_{C^m} = \max_{|\beta| \leq m} \sup_{x \in [0,1]^n} |D^\beta g(x)|.
\end{equation}

We aim to prove that under the conditions of Theorem 1, DQN with KAN as the backbone network can effectively approximate the optimal action-value function $Q^*(s,a)$. Assume the true action-value function for taking action $a$ in state $s$ is $Q^*(s,a)$. Our goal is to find an approximation function $Q(s,a;\theta)$ that is as close as possible to $Q^*(s,a)$.

Considering the mean squared error properties of DQN and $y$ given in \eqref{eq7}, we have
\begin{equation}
\mathbb{E}[(Q(s_t,\!a_t;\theta) \!-\! Q^*(s_t,\!a_t))^2] \!=\! \mathbb{E}[(Q(s_t,\!a_t;\theta) \!- y)^2 ] + C \!\!
\end{equation}
where $C = \mathbb{E}[(r_t + \gamma \max_{a'} Q(s_{t+1},a';\theta') - Q^*(s_t,a_t))^2]$ is a constant independent of $\theta$. Therefore, minimizing the loss function $\mathcal{L}(\theta)$ is equivalent to minimizing the mean squared error between the approximate value function $Q(s_t,a_t;\theta)$ and the target value $r_t+\gamma \max_{a'} Q(s_{t+1},a';\theta')$.

When we use KAN as the backbone network in DQN, the optimization objective can be rewritten as
\begin{equation}
 \min_\theta ~ \mathbb{E}[(Q(s_t,a_t;\theta) - (r_t+\gamma \max_{a'} Q(s_{t+1},a';\theta')))^2].    
\end{equation}
By using \eqref{eq12}, $Q(s_t,a_t;\theta)$ can be defined as:
\begin{align}
Q(s_t,a_t;\theta) &=\sum_{j}\sum_{i=1}^{m} (\alpha_i \cdot \text{spline}((s_t,a_t)_i;\theta_i) \notag\\
& \quad + \beta_i \cdot b((s_t,a_t)_i)) + b.
\label{eq18}
\end{align}
Since the spline functions and SiLU($x$) in~\eqref{eqsk} used in KAN are continuously differentiable, the conditions of Theorem 1 are satisfied. By applying Theorem 1, we can conclude that for any state-action pair $(s,a)$, there exists an optimal set of parameters $\theta^*$ such that $Q(s,a;\theta^*)$ in~\eqref{eq18} can arbitrarily approximate the optimal action-value function $Q^*(s,a)$.

\textit{Theorem 2}: Let $Q(s,a;\theta)$ be the approximate action-value function defined by \eqref{eq12}, where the spline functions $\text{spline}(x; \theta)$ and the activation function $b(x)$ are Lipschitz continuous with Lipschitz constants $L_{\text{spline}}$ and $L_b$, respectively. Assume that the optimal action-value function $Q^*(s,a)$ is also Lipschitz continuous with Lipschitz constant $L_{Q^*}$. Then, for any $\varepsilon > 0$, there exists a set of parameters $\theta^*$ such that
\begin{equation}
\left\lVert Q(s,a;\theta^*)-Q^*(s,a)\right\rVert_{\infty} \leq \varepsilon,
\label{eq21}
\end{equation}
and for any $\theta \in \Theta$,
\begin{equation}\label{eq2:thm2}
\left\lVert Q(s,a;\theta)-Q^*(s,a)\right\rVert_{\infty} \leq \varepsilon + C\left\lVert\theta-\theta^*\right\rVert_2
\end{equation}
where $C=\sqrt{m}(\alpha_{\max}L_{\text{spline}}+\beta_{\max}L_b)$, $m$ is the number of basis functions used in the spline approximation, $\alpha_{\max}=\max_i \alpha_i$, and $\beta_{\max}=\max_i \beta_i$.

\textit{Proof}: By the universal approximation theorem for spline functions \cite{Pillonetto2024}, $\forall \varepsilon > 0$, there is a $\theta^*$ such that
\begin{equation}
\left\lVert Q(s,a;\theta^*)-Q^*(s,a)\right\rVert_{\infty} \leq \varepsilon.
\end{equation}
For any $\theta \in \Theta$, we have
\begin{align}\label{eq1:pf thm2}
& \left\lVert Q(s,a;\theta)-Q^*(s,a)\right\rVert_{\infty}\notag \\
& \leq \left\lVert Q(s,a;\theta)-Q(s,a;\theta^*)\right\rVert_{\infty} + \left\lVert Q(s,a;\theta^*)-Q^*(s,a)\right\rVert_{\infty} \notag\\
& \leq \left\lVert Q(s,a;\theta)-Q(s,a;\theta^*)\right\rVert_{\infty} + \varepsilon.
\end{align}
By the Lipschitz continuity of $\text{spline}(x; \theta)$ and $b(x)$, we obtain
\begin{align}\label{eq2:pf thm2}
&\left\lVert Q(s,a;\theta)-Q(s,a;\theta^*)\right\rVert_{\infty} \notag\\
&\leq \sum_{j \in [1-n]} \sum_{i=1}^{m} (\alpha_i L_{\text{spline}}\left\lVert\theta_i-\theta_i^*\right\rVert_2  + \beta_i L_b \left\lVert\theta_i-\theta_i^*\right\rVert_2)  \notag\\
&\leq \sqrt{m}(\alpha_{\max}L_{\text{spline}}+\beta_{\max}L_b) \left\lVert\theta-\theta^*\right\rVert_2.
\end{align}
Combining \eqref{eq1:pf thm2} and \eqref{eq2:pf thm2} gives \eqref{eq2:thm2}. \qed

Theorem 2 provides a quantitative bound on the approximation error between the learned action-value function $Q(s,a;\theta)$ and the optimal one $Q^*(s,a)$. The bound consists of two terms: (i) The universal approximation error $\varepsilon$, which can be made arbitrarily small by choosing a suitable $\theta^*$. (ii) The term $C\left\lVert\theta-\theta^*\right\rVert_2$, which depends on the distance between the learned parameters $\theta$ and the optimal parameters $\theta^*$. The Lipschitz continuity of the spline functions and the activation function, as well as the bound on the coefficients $\alpha_i$ and $\beta_i$, ensure the stability and generalization of the learned action-value function. As the training progresses and $\theta$ approaches $\theta^*$, the approximation error decreases, indicating the convergence of the learned action-value function to the optimal one. 

Under Theorem 2, by minimizing the loss function \eqref{eq3-1}, DQN combined with KAN can effectively approximate the optimal action-value function $Q^*(s,a)$, as demonstrated by:
\begin{equation}
\lim_{\theta \to \theta^*} \mathcal{L}(\theta) \to 0 \implies \lim_{\theta \to \theta^*} Q(s,a;\theta) \to Q^*(s,a).
\end{equation}
The optimal policy $\pi^*$ selects actions maximizing the optimal Q-value $Q^*$ for each state:
\begin{equation}
\pi^*(a \mid s) := \arg \max_a Q^*(s, a).
\end{equation}
Thus, K-DQN can approximate $Q^*(s, a)$ by minimizing the loss function~\eqref{eq3-1}, enabling it to learn the optimal policy $\pi^*$. This highlights KAN’s effectiveness in enhancing DQN through direct optimization and strong theoretical guarantees.

\subsection{Computational Complexity Analysis}
Integrating KAN into DQN adds computational overhead from spline-based activations. While traditional DQN’s forward pass has complexity $O(LN^2)$ for $L$ layers and $N$ neurons, K-DQN’s complexity increases to $O(LN^2 + LNS)$, where $S$ is the number of spline segments.

During training, the backward propagation in traditional DQN has complexity $O(LN^2)$. For K-DQN, the gradient computation through spline functions adds an overhead, resulting in $O(LN^2 + LNS)$ complexity. However, two factors help mitigate this computational cost:
1) Parameter sharing among neurons reduces the number of parameters to be updated, lowering the practical computational burden to approximately $O(LN^2 + LN\bar{S})$, where $\bar{S} < S$ is the effective number of unique spline segments.
2) The improved approximation capabilities of KAN typically require fewer training iterations for convergence. Our empirical results show that K-DQN achieves equivalent performance with approximately 30\% fewer training iterations compared to traditional DQN.

During inference, with fixed spline coefficients, forward pass complexity reduces to $O(LN^2 + LN)$, causing only a slight increase in decision time over traditional DQN.



\section{Routes Planner and Actions Inspector}
\label{sec4}
This section presents mechanisms for safe and efficient roundabout driving, covering HDV driving rules, the action inspector, and the route planner.

\subsection{Driving Rules of HDVs}
This subsection outlines HDV priority rules in roundabouts to maintain traffic flow and safety, addressing common scenarios.

\subsubsection{Entry Rule}
When an HDV nears a roundabout, it must yield to vehicles already in the entrance it plans to use, ensuring smooth traffic flow and minimizing conflicts. Such a rule is described as:
\begin{equation}
\text{HDV}_{\text{entering}} \not\leftarrow \text{if } \exists \, \text{HDV}_{\text{passing}}.
\end{equation}
Inside the roundabout, ego HDVs (EHDVs) must adjust their speed per the Intelligent Driver Model (IDM) policy in \eqref{eq18:idm} to maintain a safe gap from the front HDV (FHDV) until exiting, preventing rear-end collisions and ensuring smooth flow.
\begin{equation}
 a _{\textrm{IDM}} = a_{\max} [ 1- (v_{\textrm{FHDV}} / v_{e} )^{4} - (h^{*} / h )^{2} ]
  \label{eq18:idm}
\end{equation}
where $a_{\textrm{max}}$ is the maximum acceleration of EHDV, $v_{\textrm{FHDV}}$ is the velocity of FHDV with the desired value $v_{e}$, and $h$ is the real gap between EHDV and FHDV. $h^{*}$ is the desired gap between EHDV and FHDV with the formula
\begin{equation}
h ^{*} = h_{e}+v_{\textrm{AV}}T_{e} - v_{\textrm{AV}}{\Delta v} / (2\sqrt{a_{\max} c} )
\end{equation}
where $h_{e}$ is the expected space to FHDV, $v_{\textrm{AV}}$ is AV's speed, $T_{e}$ is the desired time gap, $\Delta v$ is the velocity difference between EHDV and FHDV, and $c$ is the comfortable deceleration. 

\subsubsection{Inner Lane Following Rule}
HDVs in the inner lane of the roundabout must align their speeds with the nearest vehicle ahead, even if that vehicle is in the outer lane. This rule is intended to synchronize speeds across lanes and enhance the cohesive flow of traffic, particularly in multi-lane roundabouts. 

\subsection{Route Planner}
The integrated route planner for the ego vehicle (EV)  comprises initial-lane selection decisions, a path-planning algorithm, and a lane-change selection mechanism. The initial-lane selection is guided by the TTC metric for each lane, ensuring safety and efficiency from the start. The path planning algorithm employs a node-based shortest path calculation to determine the most optimal route. The lane-change selection mechanism is driven by a proposed lane change cost formula, facilitating effective and strategic lane changes.

\subsubsection{Initial-Lane Selection}
 By computing the TTC between the ego vehicle and surrounding vehicles, the safety levels can be ensured and unsafe actions can be avoided. In this scenario, the more potential space for driving and safety are the major considerations, thus we calculate the TTC for the inner and outer lanes as follows:
 \begin{equation}
\begin{aligned}
&\text{TTC}_{\text{inner}} = \frac{\text{Distance to HDV}_{\text{inner}}}{\text{Speed of EV} - \text{Speed of HDV}_{\text{inner}}}, \\
&\text{TTC}_{\text{outer}} = \frac{\text{Distance to HDV}_{\text{outer}}}{\text{Speed of EV} - \text{Speed of HDV}_{\text{outer}}}.
\end{aligned}
\end{equation}
The obtained TTC of both lanes can then be used to make the initial-lane selection rules. This paper considers several situations: No HDVs present, One HDV in outer lane, One HDV in both lanes, and Multiple HDVs in both lanes. These scenarios are described as follows:

$\bullet$ \textbf{No HDVs present:} The lane selection rule is 
\begin{equation}
\text{Lane}_{\text{selected}} = \text{Inner lane} \quad \text{if } \text{HDVs} = 0.
\end{equation}
With no HDVs present, the AV selects the inner lane for its shorter, more direct path through the roundabout.

\begin{figure}[t] 
    \centering
    \begin{subfigure}[b]{0.225\textwidth}
        \includegraphics[width=\linewidth]{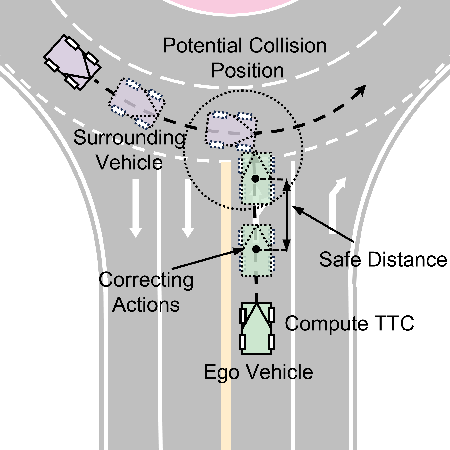}
        \vspace{-6mm}
        \caption{}
        \label{fig5:sub1}
    \end{subfigure}%
    \hspace{2mm}
    \begin{subfigure}[b]{0.225\textwidth}
        \includegraphics[width=\linewidth]{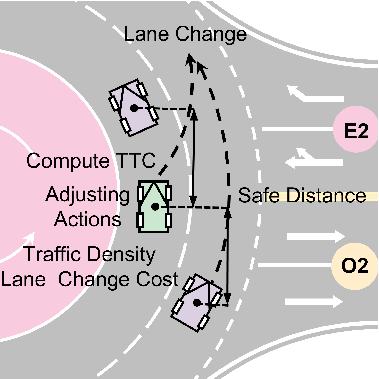}
        \caption{}
        \label{fig5:sub2}
        \end{subfigure}%
        \vspace{-1mm}
    \caption{Lane selection cases: (a) One HDV in the outer lane—EV chooses the safer lane with higher TTC; (b) One HDV in each lane—EV selects the lane with higher TTC.}
    \label{fig5}
\end{figure}

$\bullet$  \textbf{One HDV in outer lane:} Fig.~\ref{fig5}(a) illustrates this scenario, where the EV computes the TTC to maintain a safe distance from surrounding vehicles and merge into the inner lane.

$\bullet$ \textbf{One HDV in both lanes:} As illustrated in Fig.~\ref{fig5}(b), by evaluating the TTC of both lanes, the lane with the higher TTC is selected for safety and more driving space. If having equal TTC values, the inner lane is chosen to enhance efficiency. The rule is summarized as
\begin{equation}
\text{Lane}_{\text{selected}} =\! \begin{cases} 
\text{Inner lane}, &\!\! \text{if } \text{TTC}_{\text{inner}} \geq \text{TTC}_{\text{outer}} \\
\text{Outer lane}, &\!\! \text{otherwise}
\end{cases}\!. \!\!
\end{equation}
If two HDVs have the same velocity but the inner-lane HDV is farther from the EV, the inner lane is selected.

$\bullet$ \textbf{Multiple HDVs in both lanes:} This is the most complex scenario, with two HDVs in both lanes. When multiple vehicles (more than two) are present, a weighted decision based on TTC and Total Driving Time (TDT) to the outlet is applied:
\begin{equation}
\begin{aligned}
&\text{TTC}_{\text{weighted}} = w_1 \cdot \text{TTC}_{\text{nearest}} + w_2 \cdot \text{TDT},\\
&\text{TDT} = \sum \text{Driving time of each HDV to EV's outlet},
\label{eq26}
\end{aligned}
\end{equation}
where \( w_1 \) and \( w_2 \) are predefined weights reflecting traffic model preferences. The lane with the lower score is chosen to enhance safety and avoid HDV delays. The full selection process is outlined in Algorithm~\ref{alg1}.

     

\begin{algorithm}[t]
\caption{Action priority list for EV}\label{alg1}
\begin{algorithmic}[1]
\Require $L$: Lane index in the roundabout; $\alpha_1, \alpha_2, \alpha_3, \alpha_4$: Coefficients for priority score computation; 
$T_n$: Prediction horizon for trajectory planning; 
$A_{t,i}$: Feasible actions for vehicle $i$ at time $t$
\Ensure $\mathit{P_t}$: Priority list of actions for the EV.
\State Determine the presence of HDVs in the roundabout.
\If{no HDVs present}
    \State Select the inner lane.
\Else
    \State Compute TTC for each HDV in both lanes.
    \If{one HDV in each lane}
        \State Choose lane having the highest TTC value,
        \State preferring inner lane if equal.
    \ElsIf{multiple HDVs in both lanes}
        \State Use $\text{TTC}_{\text{weighted}}$ in~\eqref{eq26} to select lane.
    \EndIf
\EndIf
\State Initialize action priority list $\mathit{P_t}$ for the EV.
\For{each feasible action $a_{\text{feasible}}$ in $A_{t,i}$}
    \State Compute the priority score of $a_{\text{feasible}}$.
    \State Add $a_{\text{feasible}}$ to $\mathit{P_t}$ according to its priority score.
\EndFor
\end{algorithmic}
\end{algorithm}

\subsubsection{In-Roundabout Lane Selection}
After entering the roundabout and selecting an initial lane, the next stage is path planning.
We adopt a modified Breadth-First Search (BFS)~\cite{Chen2022} method that considers both distance and traffic conditions to compute the optimal path from a start point to a target within a graph structure, where nodes represent intersections in the road network, and edges represent drivable roads. The modified BFS algorithm uses the cost function:
\begin{equation}
C(e) = w_1 \cdot D(e) + w_2 \cdot \mathcal{D}(e)
\end{equation}
where \(C(e)\) is the cost of edge \(e\), \(D(e)\) is the distance of edge \(e\), \(\mathcal{D}(e)\) is the traffic density of edge \(e\), and \(w_1\) and \(w_2\) are weight factors that determine the relative importance of distance and traffic density. \(\mathcal{D}(e)\) is calculated by
\begin{equation}
\mathcal{D}(e) = N_e / L_e
\end{equation}
where \(N_e\) is the number of vehicles on edge \(e\), and \(L_e\) is the length of edge \(e\). The modified BFS is formulated as
\begin{equation}
\text{BFS}(s, g) = \min \{ p : s \rightarrow \ldots \rightarrow g \mid p \in \text{Paths}(s, g) \}
\end{equation}
where \(s\) is the start node, \(g\) is the goal node, and \(\text{Paths}(s, g)\) is the set of all possible paths. The optimal path is given as
\begin{equation}
p^* := \arg\min_{p \in \text{Paths}(s, g)} \sum_{e \in p} C(e)
\end{equation}
where \(p^*\) is the optimal path.
\subsubsection{Lane Selection Mechanism}

As the scenario illustrated in Fig.~\ref{fig5}(b), traffic density (\(\mathcal D \)) and lane-change cost (\(\mathcal C \)), are computed. Additionally, to align with real-world driving behaviors where vehicles preparing to exit typically move to the outer lane in advance, we implement a dynamic priority adjustment mechanism.

$\bullet$ \textbf{Traffic Density} \( \mathcal D \) is calculated by iterating over all vehicles to count the number on a specified node and lane, and adjusting the density value based on vehicles' relative positions, with closer vehicles having a higher weight. When the EV is at lane node $n$, the density is
    \begin{equation}
    \mathcal D(n, l) \!=\!\!\! \sum_{NV \in N_V}\!\! \mathbf{1}_{(NV.n = n \land NV.l = l)} - \mathbf{1}_{(NV.n = n \land NV.l \neq l)}\!\!
    \end{equation}
    where \( n \) indicates the node, \( l \) indicates the lane, \( N_V \) is the set of neighbor vehicles, and \( \mathbf{1} \) is the indicator function:
    \begin{equation}
    \mathbf{1}_{\text{condition}} = 
    \begin{cases}
    1, & \text{if the condition is true} \\
    0, & \text{otherwise}
    \end{cases}.
    \end{equation}
    
$\bullet$ \textbf{Lane Change Cost} \( \mathcal C \) is obtained by computing the distance between the controlled vehicle and other vehicles. The costs increase sharply if the distance is less than a threshold safety distance $D_{safe}$. The cost formula is
\begin{equation}
\mathcal C(n, l) = \sum_{NV \in N_V} \frac{D_{safe}}{D(EV, NV)} \cdot \mathbf{1}_{(D(EV,  NV) < D_{safe})} 
\end{equation}
where $D(EV,NV) = \| p_\textrm{EV}(t) - p_\textrm{NV}(t) \|_2$ is the distance between EV and non-ego vehicle (NV), \(p_\textrm{EV}(t)\) is EV's position, and \(p_\textrm{NV}(t)\) is NV's position. This inter-vehicle distance-based cost calculation adapts naturally to different lane geometries without parameter adjustment.

$\bullet$ \textbf{Outer Lane Preference} $\omega(d)$ is designed to encourage timely transitions to the outer lane when approaching the target exit. This preference weight is defined as:
\begin{equation}
\omega(d) = \begin{cases}
0, & \text{if } d < 0.5 d_{\text{total}} \\
\beta \cdot \frac{d - 0.5 d_{\text{total}}}{0.5 d_{\text{total}}}, & \text{otherwise} 
\end{cases}
\end{equation}
where $d$ is the distance traveled from the entrance, $d_{\text{total}}$ is the total distance to the target exit, and $\beta$ is a weighting parameter (set to 0.3 in our experiments) that controls the strength of outer lane preference. The lane choice $l_c$ is defined as:
\begin{equation}
l_c := \arg\min_{l \in \{0, 1\}} \left(\mathcal D(n, l) + \mathcal C(n, l) - \omega(d) \cdot \mathbf{1}_{(l = \text{outer})}\right).
\end{equation}
This formulation ensures that as the vehicle approaches its exit (when $d > 0.5 d_{\text{total}}$), the outer lane becomes increasingly preferable, reflecting realistic driving behavior while maintaining safety through density and lane change cost terms. When lane costs are equal, the decision is refined based on the vehicle’s position. This enhances both safety and efficiency by integrating real-time traffic conditions with potential lane change risks, while preserving natural driving tendencies. The route planner leverages a modified BFS algorithm with an edge selection function to identify optimal paths, providing a robust solution for autonomous navigation in roundabouts.

\begin{algorithm}[t]
\caption{Action execution for EV in roundabout}\label{alg2}
\begin{algorithmic}[1]
\Require $\mathit{P_t}$: Priority list of actions for the EV, initialized and populated as per the previous algorithm.
\Ensure $a_{t,i}$: The optimal action for vehicle $i$ at time step $t$, chosen and executed from the priority list.
\While{$P_t$ is not empty}
    \State $a_t \gets P_t[0]$  
    \For{$NV$ in $N_V$ and $D(EV, NV) \leq D_{safe}$}
        \If{$EV_{a_t}$ and $NV$ trajectories overlap in $T_n$}
            \If{$NV$ in same lane and in front}
                \State Use IDM~\eqref{eq18:idm} to follow FHDV; \textbf{break}
            \ElsIf{$NV$ in adjacent lane}
                \State Replace $a_t$ with the next highest priority action in $P_t$.
            \EndIf
        \EndIf
    \EndFor
    \If{no overlap}
        \State Execute $a_t$
    \EndIf
    \State Remove $a_t$ from $P_t$
\EndWhile
\end{algorithmic}
\end{algorithm}

\subsection{Action Inspector}
\label{sec4c}
Each EV plans its acceleration through the roundabout, with an incentive to accelerate for efficiency. However, to ensure safety, the EV predicts the trajectories of nearby NVs whenever their distances fall below a safety threshold $D_{safe}$, as shown in Fig.~\ref{fig5}(b). This safety distance accounts for vehicle dimensions by expanding the EV’s boundary by half of its width ($1.05\,\mathrm{m}$) and length ($2.35\,\mathrm{m}$), ensuring sufficient clearance for maneuvers. The vehicle-dimension-based safety calculation ensures robustness to varying lane widths. If predicted trajectories overlap, the EV switches to a following mode using the IDM policy~\eqref{eq18:idm}, adapting to the nearest preceding vehicle and synchronizing with traffic flow. A penalty is imposed if the EV’s speed remains below the expected value for more than three seconds, encouraging timely progression while maintaining safety.

\textit{1) Safety Margin Calculation:}
This margin is used to guide decision-making when selecting driving actions. As vehicles maintain a wider angle relative to each other while in proximity, the likelihood of their paths intersecting decreases. Therefore, the safety margin for each vehicle's maneuver is defined as the minimum difference in relative angle, $D_a$, or the shortest time until a potential collision could occur.
\begin{equation}
\label{eq:safety_margin}
\text{Safety Margin} = \min\limits_{a \in A_{\text{feasible}}} D_{\text{a},a,k}
\end{equation}
where \( A_{\text{feasible}} \) is the set of feasible actions and \( D_{\text{a},a,k} \) is the safety margin angle under action \( a \) at prediction time \( k \).

\textit{2) Decision-Making Criteria:}
For each decision point, the AV will calculate safety margins for multiple options. If the safety margins are equivalent, the AV will prefer the lane that optimizes the route, typically the inner lane in roundabouts due to the shorter path length to the exit.

\textit{3) Dangerous Action Replacement:}
When a dangerous action is detected, the inspector replaces it with the next highest-priority action from $P_t$, ensuring the EV chooses the safest option. If none are safe, the EV follows the nearest vehicle using IDM until a safe action appears.

\textit{4) Update Rule:}
After each EV action, the next highest-priority action is chosen. This repeats until the EV safely exits the roundabout, with the action inspector continuously replacing risky actions with safer ones (see Algorithm~\ref{alg2}).

The action inspector adapts by predicting $T_n$ steps ahead and continuously monitoring trajectories. For dynamic traffic density changes, the inspector updates its safety assessments at each time step using real-time traffic information. The system quickly handles unexpected HDV behaviors using priority-based action replacement and IDM following, which adapts to sudden speed changes of preceding vehicles.

\begin{algorithm}[t]
\caption{MPC controller for adjusting EV's velocity}\label{alg3}
\begin{algorithmic}[1]
\Require $v_\textrm{EV}^{*}$: The target speed of the ego vehicle.
\Ensure $u[0]$: The optimal control input for the first time step, or the output of the PID controller if no solution is found.
\State{MPC\_Controller} {$v_\textrm{EV}^{*}$}
\State $\mathit{EV} \gets \text{deepcopy}(\text{self})$
\State $\mathit{N_V} \gets \text{get\_surrounding\_vehicles}()$
\State $\mathit{opti} \gets \text{ca.Opti}()$
\State $u \gets \mathit{opti}.\text{variable}(N)$
\State $J_c \gets 0$
\For{$k \gets 0$ to $N-1$}
\For{$\text{vehicle}$ in $\mathit{NV}$}
\State $\text{action} \gets \text{use\_K-DQN}$
\State $\text{\_to\_predict\_vehicle\_action}(\text{vehicle})$
\EndFor
\State $\delta(k) \gets \text{compute\_steering}(\mathit{EV}, \mathit{NV})$
\State $\mathit{EV}.\text{update}(u(k), \delta(k))$
\State $J_c \gets J_c + (v_{\textrm{EV}}(k) - v_\textrm{EV}^{*})^2$
\For{$i \gets 1$ to $N_v$}
\State $J_c \gets J_c + (\|p_\textrm{NV}^{i}(k) - p_\textrm{EV}^{i}(k) \|_2 - D_{safe})^2$
\EndFor
\State $J_c \gets J_c + \lambda u^2(k)$
\State \text{add\_vehicle\_constraints}($\mathit{EV}$, $\mathit{NV}$, $u(k)$)
\EndFor
\State $\mathit{opti}.\text{minimize}(J_c)$
\State $\mathit{solution} \gets \mathit{opti}.\text{solve}()$
\If{$\mathit{solution}$ found}
\State \Return $\mathit{solution}.\text{value}(u[0])$
\Else
\State \Return $\text{PID\_controller}(v_{\textrm{EV}}(0), v_\textrm{EV}^{*})$
\EndIf
\end{algorithmic}
\end{algorithm}

In summary, the proposed system combines route planning and action inspection for safe, efficient AV navigation in roundabouts. TTC-based lane selection ensures safe entry, the modified BFS optimizes paths using distance and traffic data, and lane changes are guided by traffic density and cost. The action inspector monitors safety in real time, replacing risky actions to prevent collisions. By integrating global route planning with localized real-time traffic data-based lane change decisions, the proposed system demonstrates exceptional adaptability to varying traffic conditions.



\section{MPC for enhancing DRL Performance }
\label{sec5}
This section introduces the robust control for AVs including the vehicle dynamic model and MPC. 
The MPC controller considers the vehicle dynamics, collision avoidance, and other constraints in its optimization process. It predicts the future states of the EV and surrounding vehicles using the vehicle dynamic model and the actions of neighboring vehicles predicted by the DRL agent. The combination of DRL and MPC in the proposed framework brings several benefits: it allows DRL to focus on high-level decisions while MPC manages low-level controls; MPC can correct any imperfections in DRL decisions to ensure safe and feasible actions; and MPC provides a reliable, interpretable control strategy based on clear vehicle dynamics and constraints~\cite{YUAN2024111533}.

The EV's state is updated by
\begin{equation}
\begin{aligned}
p_\textrm{EV}(t+1) &= p_\textrm{EV}(t) + v_\textrm{EV}(t) \cdot \cos(h_\textrm{EV}(t)) \cdot \Delta t \\
v_\textrm{EV}(t+1) &= v_\textrm{EV}(t) + u(t) \cdot \Delta t \\
h_\textrm{EV}(t+1) &= h_\textrm{EV}(t) + v_\textrm{EV}(t) \cdot \tan(\delta(t)) \cdot \Delta t / L
\label{eq27}
\end{aligned}
\end{equation}
where \(\Delta t\) is the sampling time, \(v_\textrm{AV}(t)\) is the speed, \(h_\textrm{AV}(t)\) is the heading angle, \(L\) is the wheelbase length, $u(t)$ is the acceleration, and \(\delta(t)\) is the steering angle.

The following control input and collision avoidance constraints are applied to ensure safety and feasibility:
\begin{equation}\label{MPC constraints}
\begin{aligned}
v_{\min} & \leq v_\textrm{EV}(t)\leq  v_{\max},~a_{\textrm{min}} \leq u(t) \leq a_{\textrm{max}},  \\
  \delta_{\textrm{min}} &\leq \delta(t) \leq \delta_{\textrm{max}},~ \| p_\textrm{EV}(t) - p_\textrm{NV}(t) \|_2  \geq D_{safe}.
\end{aligned}
\end{equation}

At time step $t$, the optimal solutions $u^{*}(t)$ and $\delta^{*}(t)$ are obtained by solving the optimization problem:
\begin{align}
& \hspace{5em} \min J_c \\
\text{s.t.}~& \eqref{eq27}, \eqref{MPC constraints}, NV \in N_V, k \in [0, N_p-1] \notag
\end{align}
with the cost function $J_c=  \sum_{k=0}^{N_p-1} (v_\textrm{AV}(k) - v_\textrm{AV}^{*})^2 + \sum_{k=0}^{N_p-1} \sum_{i=1}^{N_v}(\lVert p_\textrm{SV}^{i}(k) - p_\textrm{AV}^{i}(k) \rVert - D_{safe})^2  + \lambda \sum_{k=0}^{N_c-1} u^2(k)$.
$N_p$ and $N_c$ represent the prediction horizon and control horizon, respectively. \(v_\textrm{AV}^{*}(k)\) is the target speed, and \(\lambda\) is a given weighting factor. In our experiments, we set $N_p$ as 10 and the $N_c$ as 5. The entire control process is summarized in Algorithm~\ref{alg3}.

\section{Simulation Results}
\label{sec6}

\begin{figure}[t] 
    \centering
    \begin{subfigure}[b]{0.24\textwidth}
        \includegraphics[width=\linewidth]{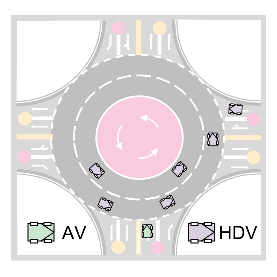}
        \vspace{-7mm}
        \caption{Normal mode settings}
        \label{fig7:sub1}
    \end{subfigure}%
     \hspace{0pt}
    \begin{subfigure}[b]{0.24\textwidth}
        \includegraphics[width=\linewidth]{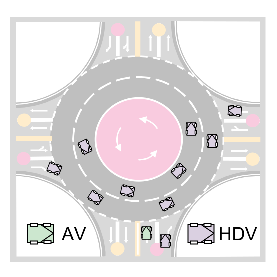}
        \vspace{-7mm}
        \caption{Hard mode settings}
        \label{fig7:sub2}
    \end{subfigure}%
    \caption{Validation settings: (a) Normal mode with six initial HDVs; (b) Hard mode with ten HDVs.}
    \label{fig7}
\end{figure}

\begin{figure*}[t] 
    \centering
    \begin{subfigure}[b]{0.32\textwidth}
        \includegraphics[width=\linewidth]{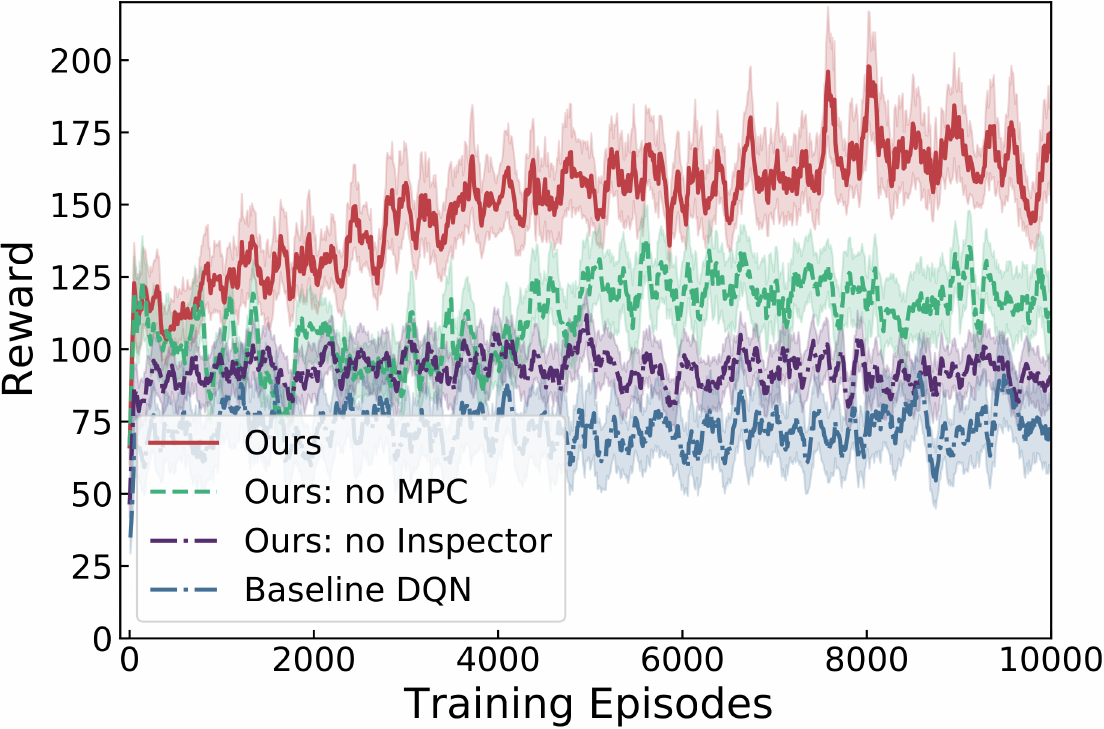}
        \caption{}
        \label{fig6:sub1}
    \end{subfigure}%
     \hspace{5pt}
    \begin{subfigure}[b]{0.315\textwidth}
        \includegraphics[width=\linewidth]{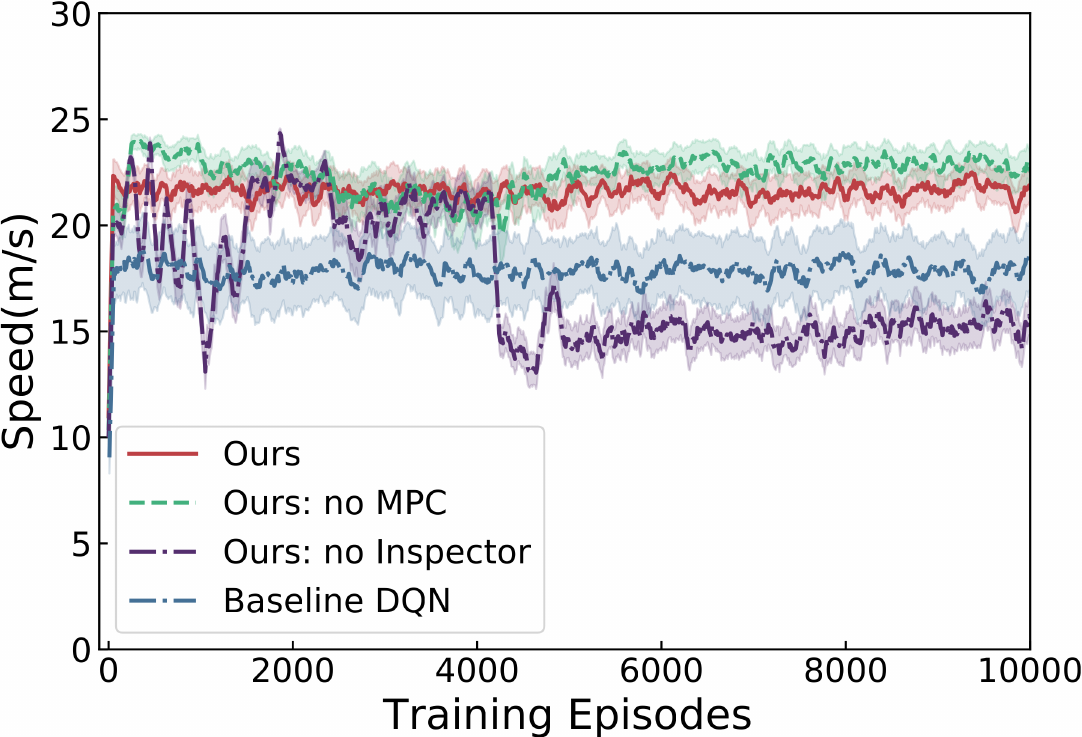}
        \caption{}
        \label{fig6:sub2}
    \end{subfigure}%
    \hspace{5pt}
    \begin{subfigure}[b]{0.325\textwidth}
     \renewcommand\thesubfigure{\alph{subfigure}}
        \includegraphics[width=\linewidth]{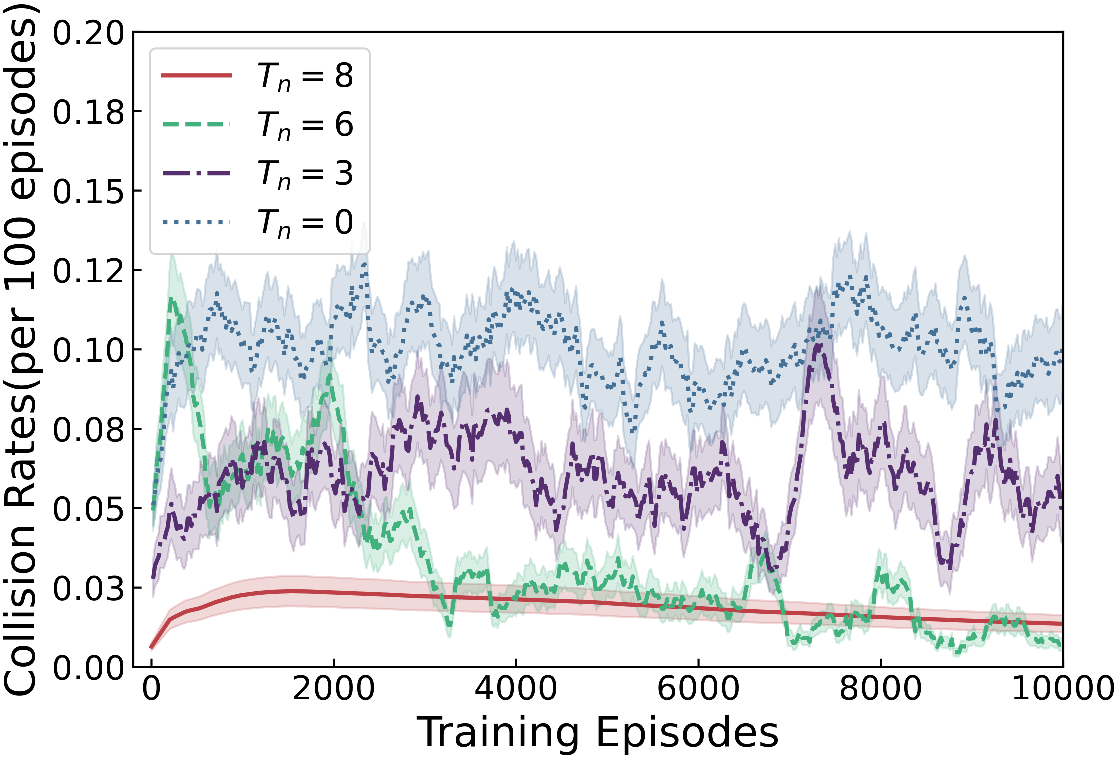}
       \caption{}
        \label{fig6:sub3}
    \end{subfigure}%
    \vspace{-2mm}
\caption{Performance comparison with different components and prediction steps. (a) reward, (b) speed, and (c) collision rates under different prediction steps ($T_n$). The shaded regions denote the standard deviations over 3 random seeds.} All curves are smoothed over the last 9 evaluation episodes.
    \label{fig6}
\end{figure*}

We evaluate K-DQN in the roundabout scenario described in Section~\ref{sec2}, focusing on training efficiency and collision rate. All experiments are conducted using three random seeds, with mean results plotted and standard deviations shown as shaded areas. The roundabout follows standard traffic engineering design, with an inner radius of $20\,\mathrm{m}$, outer radius of $28\,\mathrm{m}$, and a lane width of $4\,\mathrm{m}$. All vehicles are modeled as typical passenger cars ($4.7\,\mathrm{m}$ long, $2.1\,\mathrm{m}$ wide), and these dimensions are considered in collision detection and safety distance calculations. The algorithm employs a relative state representation and distance-based safety mechanism, ensuring robustness to varying lane widths. Vehicles exiting the roundabout leave the AV’s observation range but continue updating their kinematics. We examine three scenarios as follows:

\begin{itemize}
\item \textit{Ablation study in hard mode:} The proposed K-DQN is compared with K-DQN without the action inspector, K-DQN without MPC, and the baseline DQN.
\item \textit{Normal mode validation:} The proposed K-DQN is compared with benchmarks with seven initial vehicles in the roundabout as depicted in Fig.~\ref{fig7}(a).
\item \textit{Hard mode validation:} The proposed K-DQN is compared with benchmarks with eleven initial vehicles in the roundabout as depicted in Fig.~\ref{fig7}(b).
\end{itemize}

The benchmarks used for comparison in the normal and hard mode validations include PPO~\cite{9693175}, A2C~\cite{Hou2024}, ACKTR~\cite{Zhu2024}, and DQN~\cite{cai2022dq}. The performance metrics used for evaluation include training convergence rate, collision rate, average speed, and reward values during training and evaluation.
Considering the inherent risks associated with real-world vehicles and the constraints imposed by legal regulations, scenario-based virtual testing offers significant benefits like precise environmental replication and enhanced testing efficiency. Therefore, this study employs scenarios developed on the Highway virtual simulation platform~\cite{highway-env}. To ensure robust performance under varying traffic conditions, HDV speeds are initialized following a normal distribution around a mean speed of $20\,\mathrm{m/s}$ with a $\pm$15\% variation. Similarly, HDV's positions are randomized with a normal distribution to create diverse traffic patterns. At the end of each episode, the vehicles and their velocities are slightly randomized at their spawn points to enhance the generalization capability of our proposed model. This randomization helps evaluate the model's adaptability to different traffic speeds and densities, which is crucial for real-world deployment. The computer and environment setup for this study include Python 3.6, PyTorch 1.10.0, Ubuntu 20.04.6 LTS OS, a Intel\textsuperscript{\textregistered} Core\texttrademark\ i5-12600KF CPU, an NVIDIA GeForce RTX 3090 GPU, and 64GB of RAM.

\subsection{Ablation Study in Hard Mode}
This section describes experiments to evaluate the crucial functions of the action inspector and MPC of the proposed system in hard mode. To comprehensively assess the contribution of each component, we include a baseline DQN (using traditional MLP architecture) for comparison with our K-DQN variants. We divide the experiments into four validations: Validation 1 evaluates training performance of K-DQN, K-DQN without action inspector, K-DQN without MPC, and baseline DQN. Validation 2 tests the stability of the speed variations. Validation 3 compares the reward across the evaluation. Validation 4 analyzes the number of collisions.

\begin{figure}[t]
    \centering
    \includegraphics[width=0.7\linewidth]{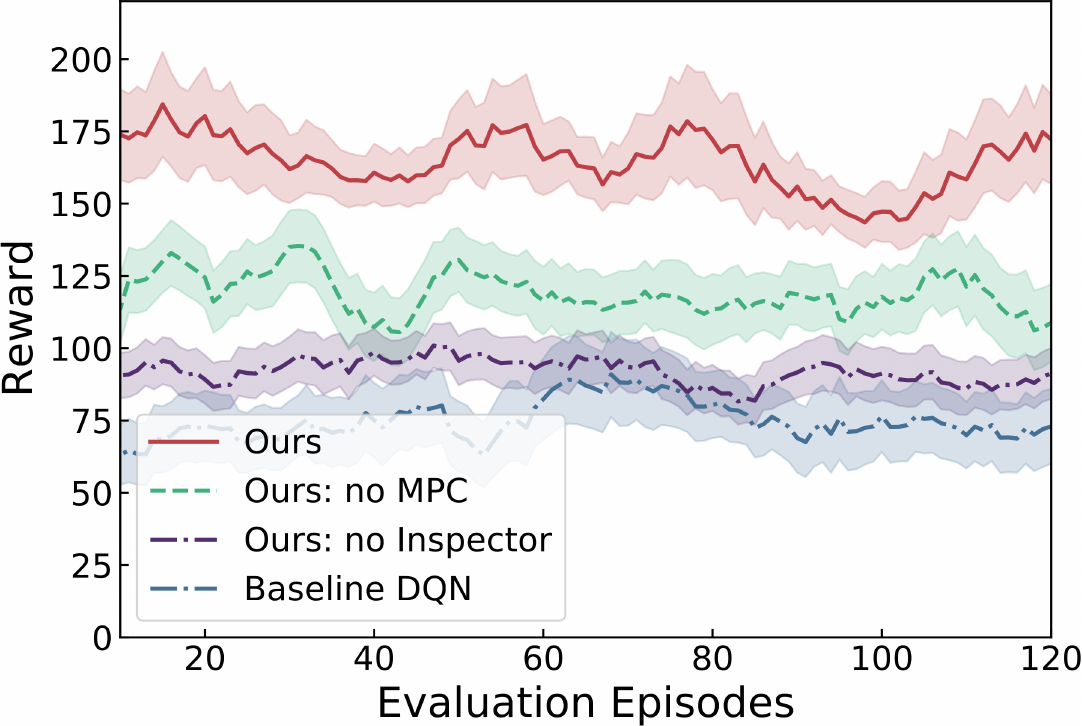}
    \caption{Rewards of different K-DQN schemes and DQN.}
    \label{fig10}
\end{figure}

\begin{figure*}[t] 
    \centering
    \begin{subfigure}[b]{0.32\textwidth}
        \includegraphics[width=\linewidth]{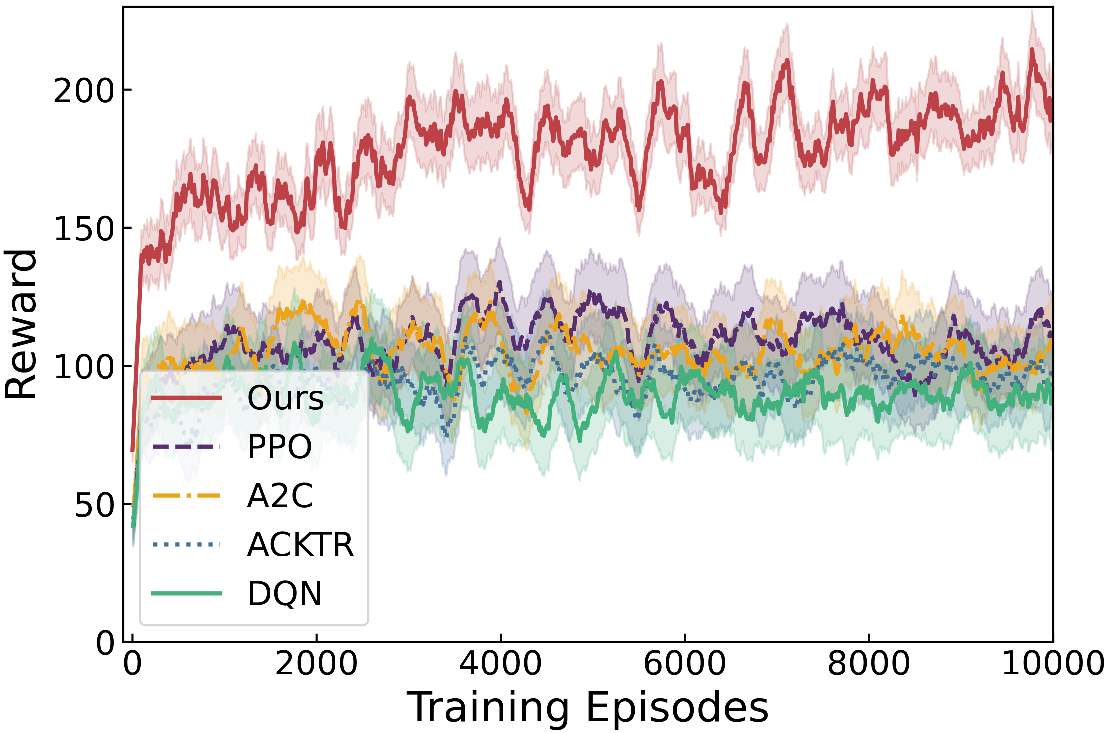}
        \caption{}
        \label{fig11:sub1}
    \end{subfigure}%
     \hspace{5pt}
    \begin{subfigure}[b]{0.32\textwidth}
        \includegraphics[width=\linewidth]{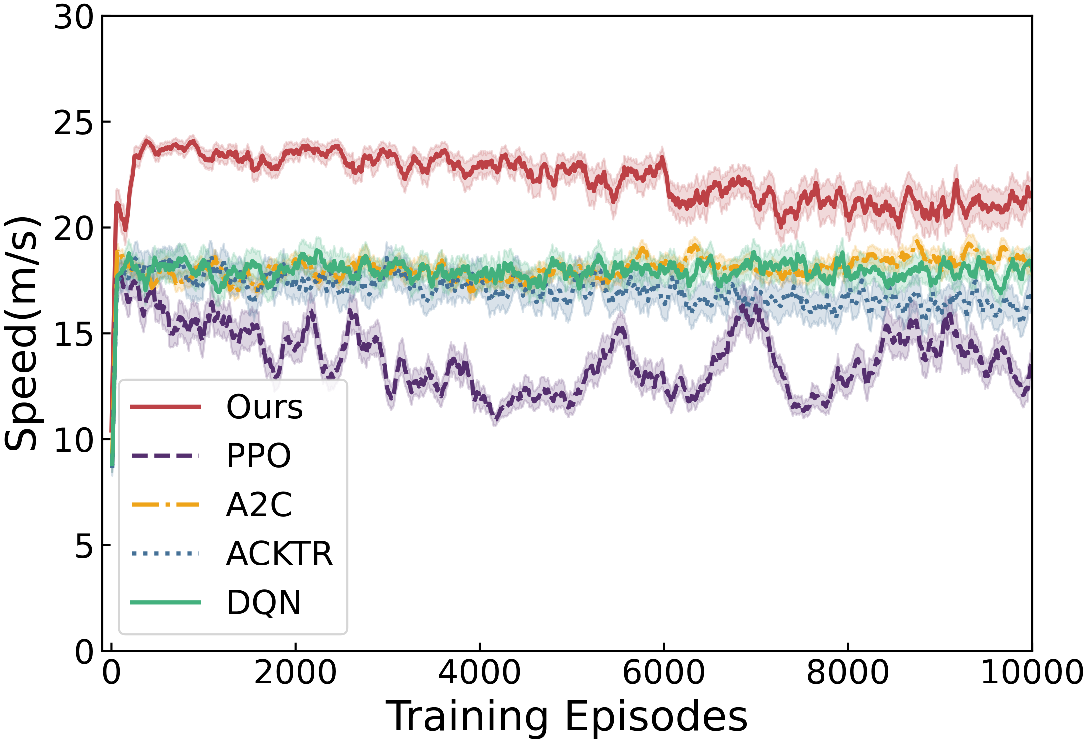}
        \caption{}
        \label{fig11:sub2}
    \end{subfigure}%
    \hspace{5pt}
    \begin{subfigure}[b]{0.32\textwidth}
        \includegraphics[width=\linewidth]{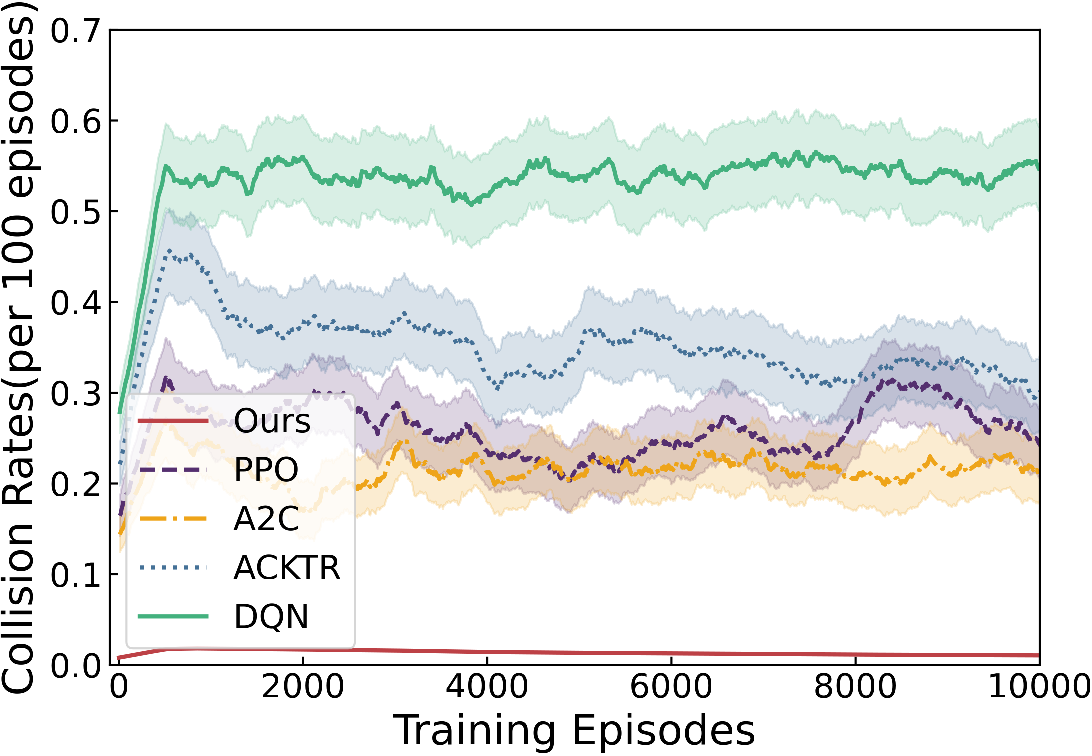}
        \caption{}
        \label{fig11:sub3}
    \end{subfigure}%
    \vspace{-2mm}
    \caption{Performance comparison with benchmarks in normal mode: (a) reward, (b) speed, and (c) collision rate.}
    \label{fig11}
\end{figure*}

\textbf{Validation 1: Training performance.}
To better assess performance under random HDV behavior, we conducted tests using three random seeds and varied scenarios. Figure~\ref{fig6}(a) compares training curves of the proposed K-DQN with three variants: K-DQN without action inspector, K-DQN without MPC, and baseline DQN with a standard MLP architecture. As expected, the full K-DQN consistently achieves higher peak rewards and faster, more stable convergence. Compared to baseline DQN, the K-DQN without inspector already benefits from the KAN architecture, yielding 10–15\% higher rewards. However, the largest gain comes from the action inspector, which filters unsafe actions and reduces collision penalties during exploration. The proposed K-DQN shows low variance across seeds, indicating stable training, while K-DQN without the inspector or MPC suffers from more fluctuations and slower convergence. The baseline DQN performs the worst, confirming that both the KAN architecture and action inspector are essential for robust and efficient learning.

\begin{table}[t]
\centering
\setlength{\tabcolsep}{5pt}
\captionsetup{
   labelfont={sc}, 
   textfont={sc}, 
   labelsep=colon, 
   skip=1em,     
   singlelinecheck=false,
   justification=centering,
   format=plain
}
\caption{Collision Rates and Average Speeds for Different K-DQN schemes}
\label{tab3}
\begin{threeparttable}
\begin{tabular}{c|c|c|c}
\hline\hline
Metrics & No MPC & No Inspector & Ours \\
\hline
Collision Rate (\%) & 9 & 11 & \textbf{2} \\
\hline
Average Speed (m/s) & \textbf{22.88} & 16.23 & 22.37 \\
\hline\hline
\end{tabular}
\begin{tablenotes}
\item Collision rate is measured per 100 episodes during training. The best results are highlighted in bold.
\end{tablenotes}
\end{threeparttable}
\end{table}

\textbf{Validation 2: Stability of the speeds variation.} Figure~\ref{fig6}(b) compares the speed profiles across different configurations: the proposed K-DQN, K-DQN without action inspector, K-DQN without MPC, and baseline DQN. The baseline DQN maintains relatively stable yet suboptimal speeds around $17$–$18\,\mathrm{m/s}$, reflecting a conservative strategy due to $\epsilon$-greedy exploration without safety constraints, which often leads to aggressive or overly cautious decisions in dense traffic~\cite{Li2024}. K-DQN without inspector exhibits significant early fluctuations (episodes 0–4000), as the KAN architecture aggressively explores without safety filtering, eventually stabilizing at a lower speed of about $15\,\mathrm{m/s}$ to mitigate collision risk. This shows that while KAN improves representation, it cannot alone resolve the safety-efficiency trade-off. In contrast, the full K-DQN achieves the best balance, maintaining higher speeds near $22\,\mathrm{m/s}$ with minimal variation ($<2\,\mathrm{m/s}$), while K-DQN without MPC shows larger fluctuations around $4\,\mathrm{m/s}$. These results highlight the complementary roles of the action inspector, which enables safe high-speed exploration, and MPC, which ensures stable policy execution. This does not undermine the value of DQN as a foundation; rather, it underscores the effectiveness of our enhancements. The KAN architecture enhances function approximation for safer policy learning, the action inspector reduces collisions dramatically (from 52\% to 2\%) by filtering unsafe actions, and MPC guarantees smooth, reliable execution. Together, these components enable DQN-based systems to reach state-of-the-art performance in safety-critical autonomous driving scenarios.

Figure~\ref{fig6}(c) shows how prediction horizon ($T_n$) affects the action inspector. With $T_n=8$, the collision rate is lowest and most stable ($\sim 0.02$), allowing ample time for action replacement. Shorter horizons ($T_n=6$ and 3) raise the rate to $\sim 0.03$ and $\sim 0.07$, while $T_n=0$ performs worst ($\sim 0.11$), highlighting the importance of predictive action inspection.

\textbf{Validation 3: Reward values among evaluation.} Figure~\ref{fig10} compares rewards for K-DQN, its ablated variants, and baseline DQN. K-DQN achieves the highest average ($\sim 175$), showing the benefit of integrating the action inspector and MPC. Removing MPC drops rewards to $\sim 125$, and without the inspector, rewards fall below 100 due to frequent collisions. Lastly, the baseline DQN performs the worst, with rewards mostly staying below 75, underscoring the limitations of naive reinforcement learning without structured safety or strategic reasoning. These results collectively highlight the necessity of combining multi-level safety and planning mechanisms for robust autonomous decision-making.

\begin{figure}[t]
    \centering
    \includegraphics[width=0.7\linewidth]{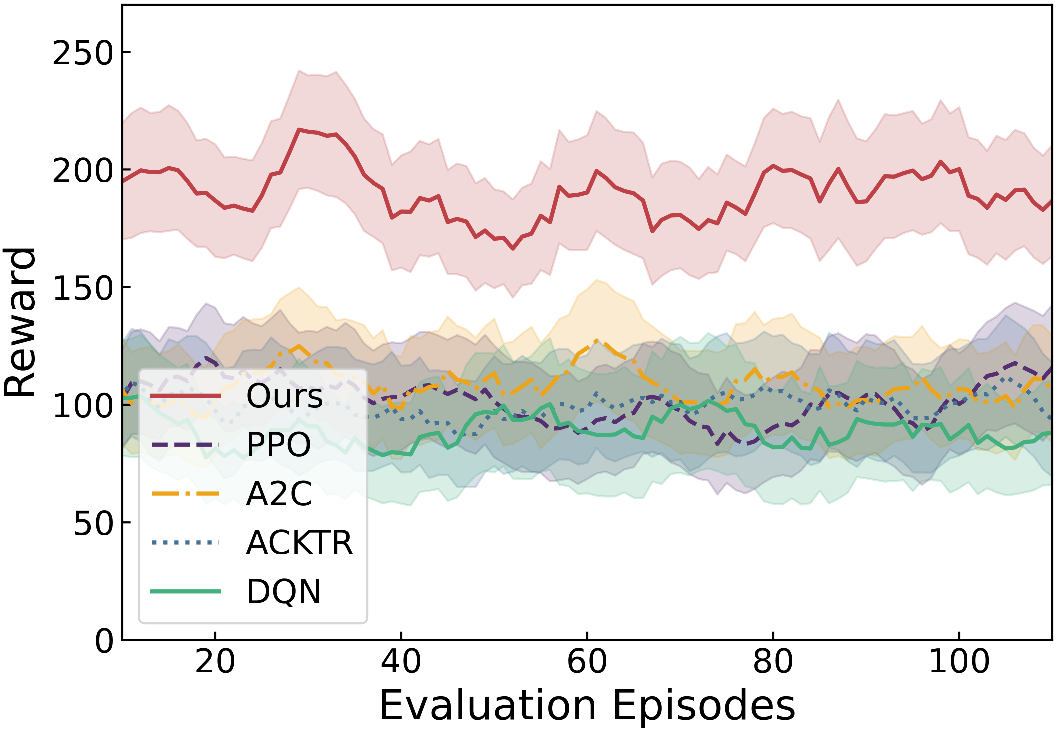}
    \caption{Rewards for K-DQN and benchmarks: normal mode.}
    \label{fig12}
\end{figure}

\textbf{Validation 4: Collision rate.}
Table~\ref{tab3} compares collisions and average speed for the proposed K-DQN, K-DQN without the action inspector, and K-DQN without MPC. The full K-DQN achieves the lowest collision rate at $2\%$ with a solid average speed of $16.23~\mathrm{m/s}$. Without the action inspector, speed drops and collisions rise to $11\%$. Removing MPC increases speed to $22.88 ~\mathrm{m/s}$, but the collision rate remains high at $9\%$. These results highlight the full K-DQN's strong balance of safety and efficiency.

\subsection{Normal Mode Validation}
This section presents the experiments in normal mode (with seven initial vehicles in the roundabout in Fig.~\ref{fig7}(a)) with comparison to benchmark DRL algorithms, PPO~\cite{9693175}, A2C~\cite{Hou2024}, ACKTR~\cite{Zhu2024}, and DQN~\cite{cai2022dq}. 

\begin{figure*}[t] 
    \centering
    \begin{subfigure}[b]{0.32\textwidth}
        \includegraphics[width=\linewidth]{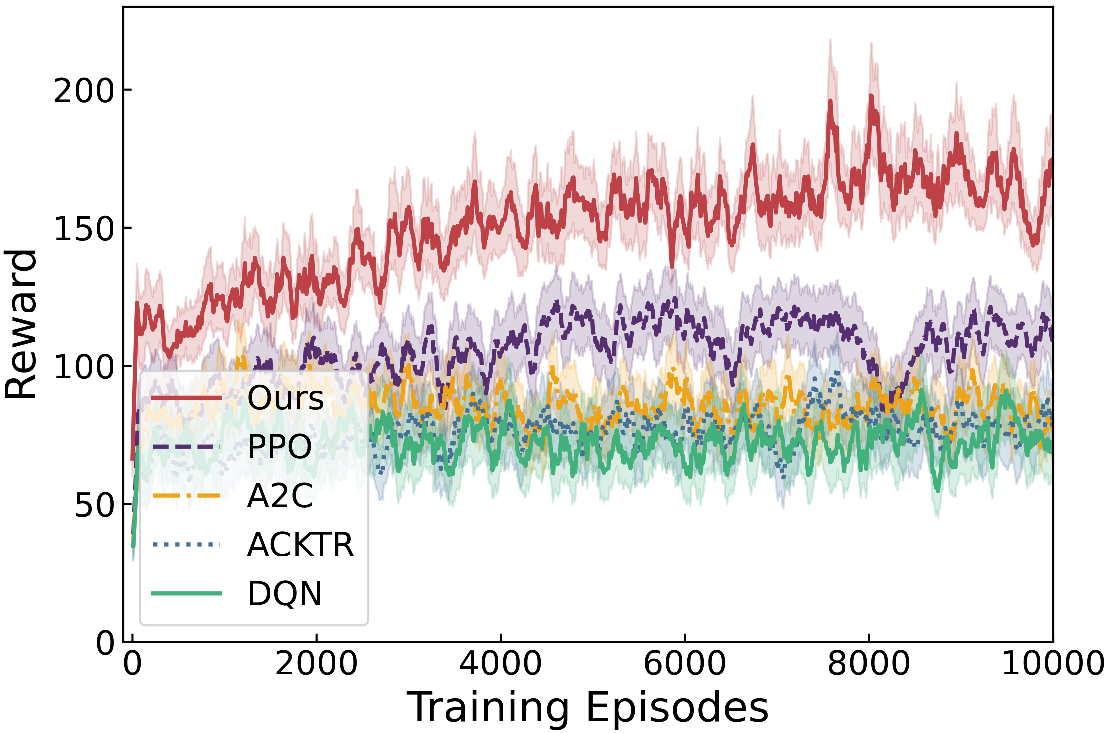}
        \caption{}
        \label{fig:sub1}
    \end{subfigure}%
     \hspace{5pt}
    \begin{subfigure}[b]{0.32\textwidth}
        \includegraphics[width=\linewidth]{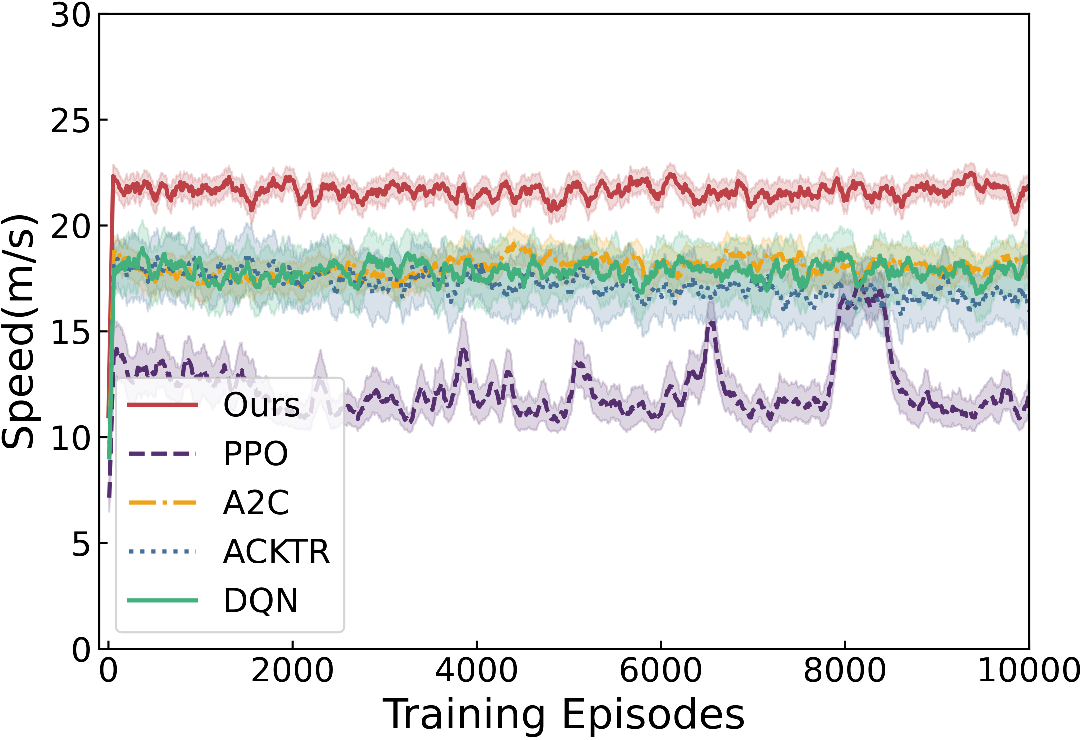}
        \caption{}
        \label{fig:sub2}
    \end{subfigure}%
    \hspace{5pt}
    \begin{subfigure}[b]{0.32\textwidth}
        \includegraphics[width=\linewidth]{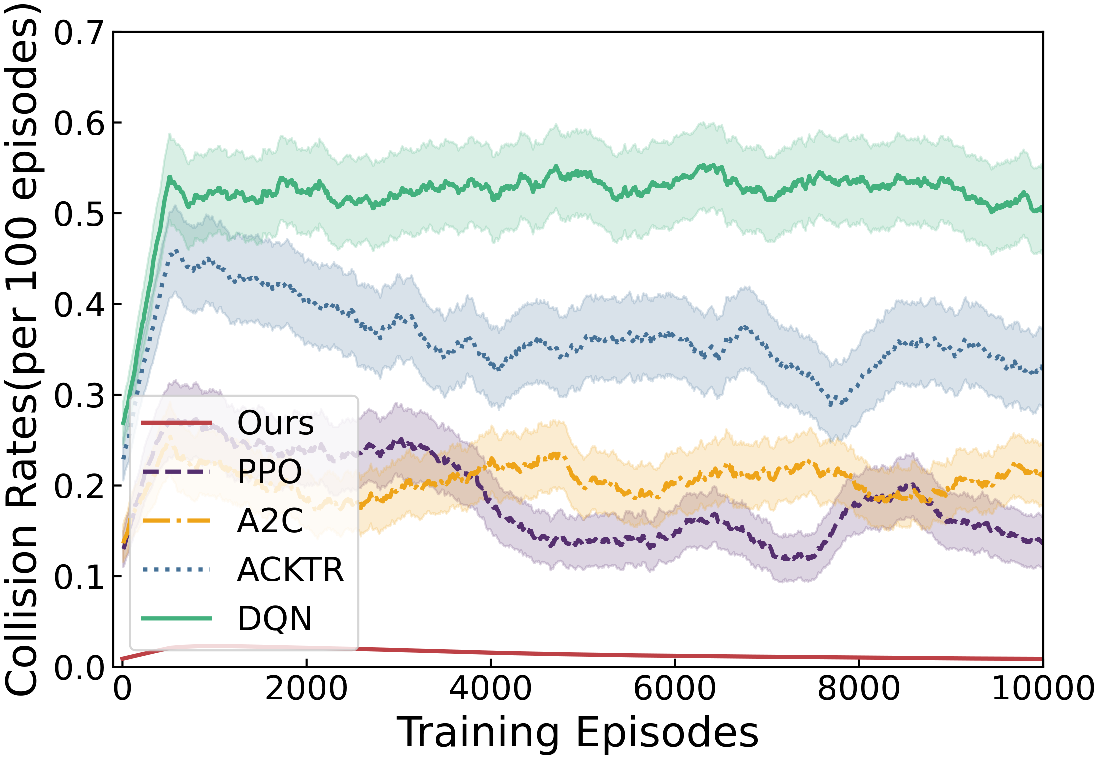}
        \caption{}
        \label{fig:sub3}
    \end{subfigure}
    \vspace{-2mm}
    \caption{Performance comparison with benchmarks in hard mode: (a) reward, (b) speed, and (c) collision rate.}
    \label{fig13}
\end{figure*}

\textbf{Validation 1: Training performance.}
To assess training performance, we test K-DQN and benchmarks using three random seeds and varied scenarios.
Figure~\ref{fig11}(a) shows K-DQN achieves the highest rewards (over 200) and fastest convergence, outperforming PPO ($\sim 125$), A2C and ACKTR (similar), and DQN ($\sim 80$).
Figure~\ref{fig11}(b) shows K-DQN maintains the highest and most stable speed ($22~\mathrm{m/s}$), while PPO has the lowest and most unstable. A2C reaches ~$18~\mathrm{m/s}$; others are similar.
Figure~\ref{fig11}(c) shows K-DQN achieves the lowest collision rate ($<0.05$), compared to PPO and A2C ($\sim 0.2$), ACKTR ($\sim 0.35$), and DQN ($\sim 0.55$).

\textbf{Validation 2: Reward values among evaluation.} Figure~\ref{fig12} illustrates the reward comparison between the proposed K-DQN and other benchmark algorithms. DQN has the lowest reward during the evaluation, falling below 75. A2C and ACKTR are similar, both increasing the reward to around 75. PPO has a relatively higher reward of around 100, peaking at 125. The reward of the proposed K-DQN fluctuates around 175, significantly surpassing other benchmark algorithms.

\subsection{Hard Mode Validation}
This validation assesses the proposed system’s safety and efficiency in the most challenging conditions, with eleven initial vehicles in the roundabout as depicted in Fig.~\ref{fig7}(b).


\begin{figure}[t]
    \centering
    \includegraphics[width=0.7\linewidth]{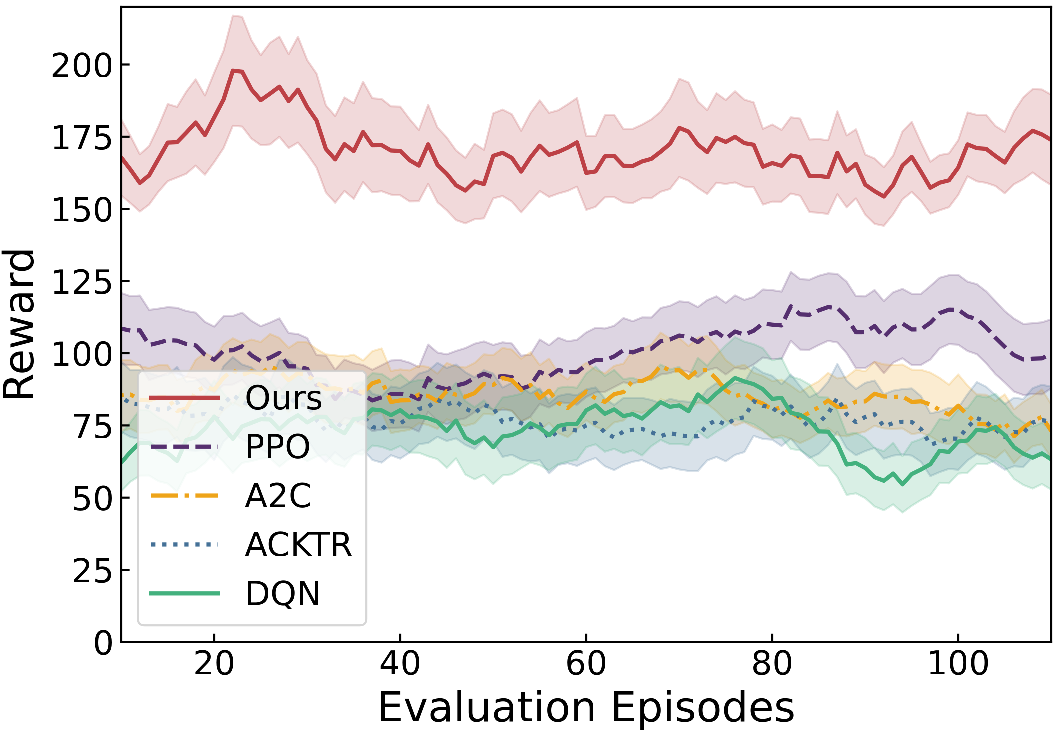}
    \vspace{-1.5mm}
    \caption{Rewards for K-DQN and benchmarks: hard mode.}
    \label{fig14}
\end{figure}

\textbf{Validation 1: Training performance.}
To ensure a robust evaluation, three random seeds are used to generate diverse scenarios for comparison with benchmark algorithms. 
Figure~\ref{fig13}(a) shows K-DQN achieved the highest peak rewards ($\sim 160$) and the fastest convergence, outperforming the closest competitor ($\sim 100$) and significantly surpassing DQN (75). In Fig.\ref{fig13}(b), K-DQN maintains higher and more stable speeds, peaking at $2~ \mathrm{m/s}$, while PPO shows the lowest and most unstable performance ($11$–$16~\mathrm{m/s}$). A2C, ACKTR, and DQN follows similar trends, with A2C reaching up to $18 ~\mathrm{m/s}$. Figure~\ref{fig13}(c) highlights K-DQN’s superior safety, with a collision rate below 0.05, compared to DQN (0.52), ACKTR (0.35), and the more variable rates of A2C ($\sim 0.2$) and PPO ($\sim 0.15$). These results confirm K-DQN's strong training efficiency, stability, and safety, supporting its potential for real-world deployment.

\textbf{Validation 2: Reward values among evaluation.}
The evaluation, using three random seeds to ensure scenario diversity (Fig.~\ref{fig14}), highlights the K-DQN algorithm’s superior reward performance. Traditional DQN remained below 70, while A2C and ACKTR show slight improvements near 75. PPO reaches a peak of 125 but averaged around 100. In contrast, K-DQN consistently outperforms all baselines, maintaining reward levels around 175, demonstrating clear dominance in reward maximization.

\begin{table}[t]
\centering
\setlength{\tabcolsep}{5pt}
\captionsetup{
    labelfont={sc}, 
    textfont={sc}, 
    labelsep=colon, 
    skip=1em,     
    singlelinecheck=false,
    justification=centering,
    format=plain
}
\caption{Collision Rates and Average Speeds for the Proposed Method and Benchmarks}
\label{tab4}
\begin{threeparttable}
\begin{tabular}{c|c|c|c|c|c|c}
\hline\hline
Scenarios & Metrics & PPO & A2C & ACKTR & DQN & Ours \\
\hline
\multirow{2}{*}{\parbox{1cm}{ Normal\\ Mode}} & coll. rate (\%) & 23 & 21 & 28 & 52 & \textbf{1} \\
\cline{2-7}
& avg. v (m/s) & 14.76 & 18.83 & 17.89 & 18.31 & \textbf{21.59} \\
\hline
\multirow{2}{*}{\parbox{1cm}{\raggedright Hard Mode}} & coll. rate (\%) & 12 & 19 & 31 & 52 & \textbf{2} \\
\cline{2-7}
& avg. v (m/s) & 13.67 & 18.04 & 17.53 & 17.70 & \textbf{22.52} \\
\hline\hline
\end{tabular}

\begin{tablenotes}
\item coll. rate means collision rate per 100 episodes during training. The best results are highlighted in bold.
\end{tablenotes}
\end{threeparttable}

\end{table}

Table~\ref{tab4} compares collision rates and average speeds of the proposed K-DQN against PPO, A2C, ACKTR, and DQN in both normal and hard modes. In normal mode, K-DQN achieves the lowest collision rate (0.01), outperforming PPO (0.23), A2C (0.21), ACKTR (0.28), and DQN (0.52). It also records the highest average speed at $21.59~\mathrm{m/s}$, surpassing PPO ($14.76~\mathrm{m/s}$), A2C ($18.83~\mathrm{m/s}$), ACKTR ($17.89~\mathrm{m/s}$), and DQN ($18.31~\mathrm{m/s}$). In hard mode, K-DQN maintains its advantage with the lowest collision rate (0.02) and highest speed ($22.52~\mathrm{m/s}$), while all benchmarks show reduced safety and efficiency. These results confirm K-DQN’s superior safety, efficiency, and robustness across varying traffic complexities.

This study employs average vehicle speed as a key performance metric, reflecting roundabout capacity and traffic efficiency. Higher average speeds imply improved flow and increased capacity. Comparative analysis shows that the proposed K-DQN consistently outperforms benchmark algorithms in both safety and efficiency. While PPO exhibits moderate performance, it lags behind K-DQN in both metrics. A2C and ACKTR perform better than PPO and DQN but fall short of K-DQN. DQN records the highest collision rate, despite maintaining relatively high speeds. Overall, K-DQN achieves lower collision rates and higher average speeds across both normal and hard scenarios, highlighting its effectiveness for safe, efficient navigation in complex traffic environments.

\section{Conclusion}
\label{sec7}
This paper proposes a DRL-based algorithm to improve AV safety and efficiency in complex roundabout traffic with HDVs. Using a DQN that processes surrounding vehicle states, it avoids manual feature engineering and enhances environmental perception. The integration of a KAN further improves learning accuracy and reliability. The algorithm includes an action inspector to reduce collisions, a route planner for efficient driving, and MPC control for stable, precise actions. Evaluations show superior performance with fewer collisions, reduced travel times, and faster training convergence compared to state-of-the-art benchmarks. Future research will focus on: 1) evaluating the algorithm’s robustness in more complex traffic scenarios, including urban ramps with elevation changes, multi-lane intersections with pedestrian crossings, and roundabouts with varied lane widths ($3.5\text{--}5.0~\mathrm{m}$) and radii to assess adaptability to diverse infrastructures; 2) enhancing driving strategies by incorporating passenger comfort (jerk minimization) and energy efficiency for electric vehicles; and 3) extending the algorithm to collaborative multi-agent settings with V2V communication and platooning to evaluate performance in cooperative and competitive urban traffic.

\bibliographystyle{IEEEtran}
\bibliography{IEEEabrv,zq_lib}

\vfill
\begin{IEEEbiography}[{\includegraphics[width=1in,height=1.25in,clip,keepaspectratio]{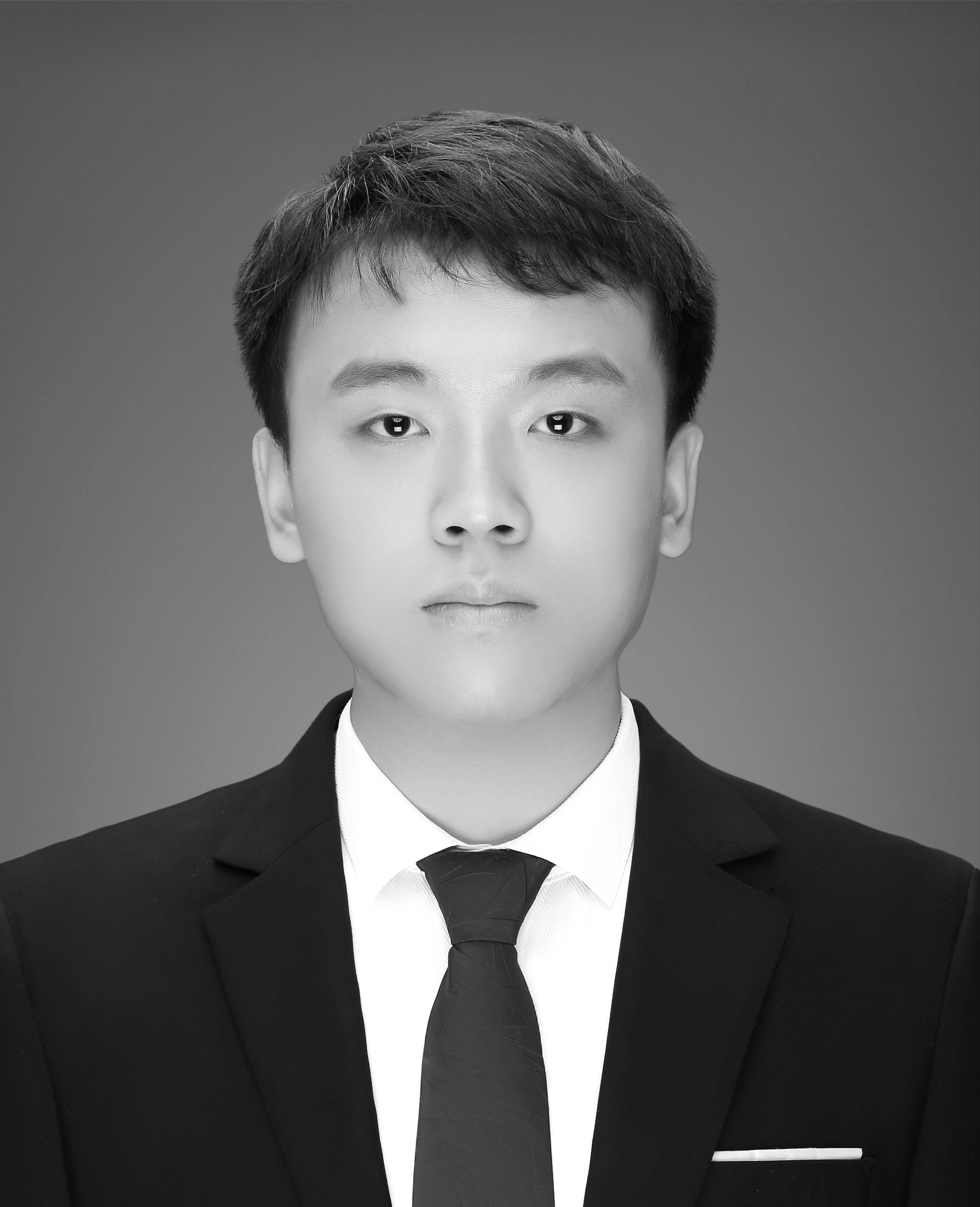}}]{Zhihao Lin}received an M.S. degree from the College of Electronic Science \& Engineering, Jilin University, Jilin, China. He is currently pursuing a Ph.D. degree with the College of Science and Engineering, University of Glasgow, Glasgow, U.K. His main research interests focus on multi-sensor fusion SLAM systems, reinforcement learning, and hybrid control of vehicle platoons.
\vspace{-5mm}
\end{IEEEbiography}

\begin{IEEEbiography}[{\includegraphics[width=1in,height=1.25in,clip]{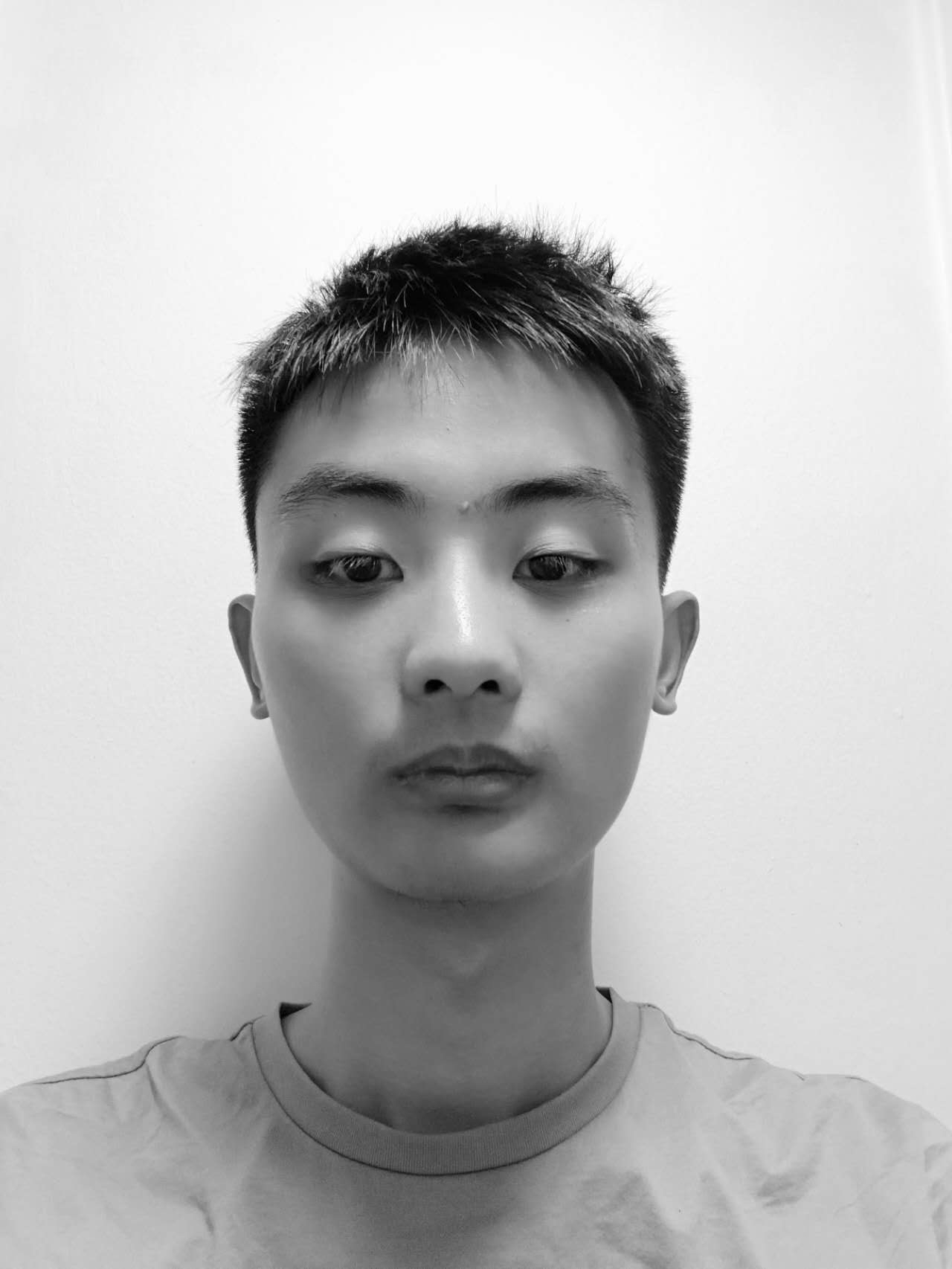}}]{Zhen Tian}received his B.S. degree in electronic and electrical engineering from the University of Strathclyde, Glasgow, U.K. He is currently pursuing a Ph.D. degree with the College of Science and Engineering, University of Glasgow, Glasgow, U.K. His main research interests include interactive vehicle decision systems and autonomous racing decision systems.
\vspace{-5mm}
\end{IEEEbiography}

\begin{IEEEbiography}
[{\includegraphics[width=1in,height=1.25in, clip]{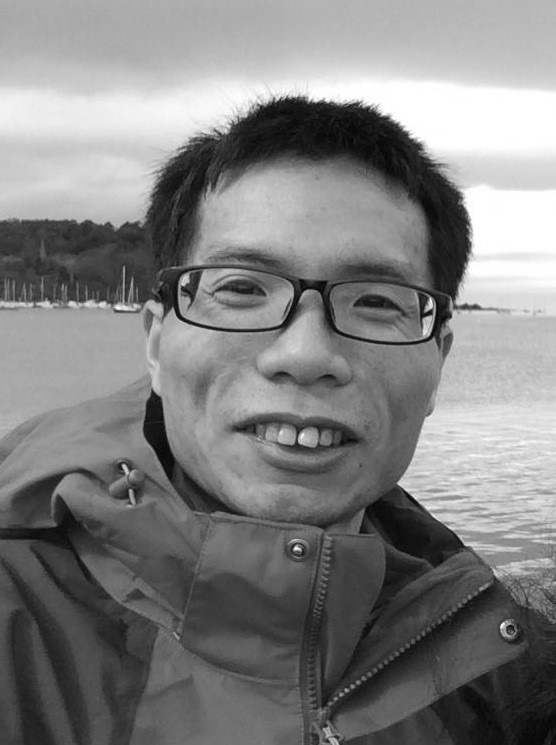}}]{Jianglin Lan}received a Ph.D. degree from the University of Hull in 2017. He has been a Leverhulme Early Career Fellow and Lecturer at the University of Glasgow since 2022. He was a Visiting Professor at the Robotics Institute, Carnegie Mellon University, in 2023. From 2017 to 2022, he held postdoc positions at Imperial College London, Loughborough University, and University of Sheffield. His research interests include safe AI, fault-tolerant systems, autonomous vehicles, and robotics.
\vspace{-5 mm}
\end{IEEEbiography}

\begin{IEEEbiography}
[{\includegraphics[width=1in,height=1.25in, clip]{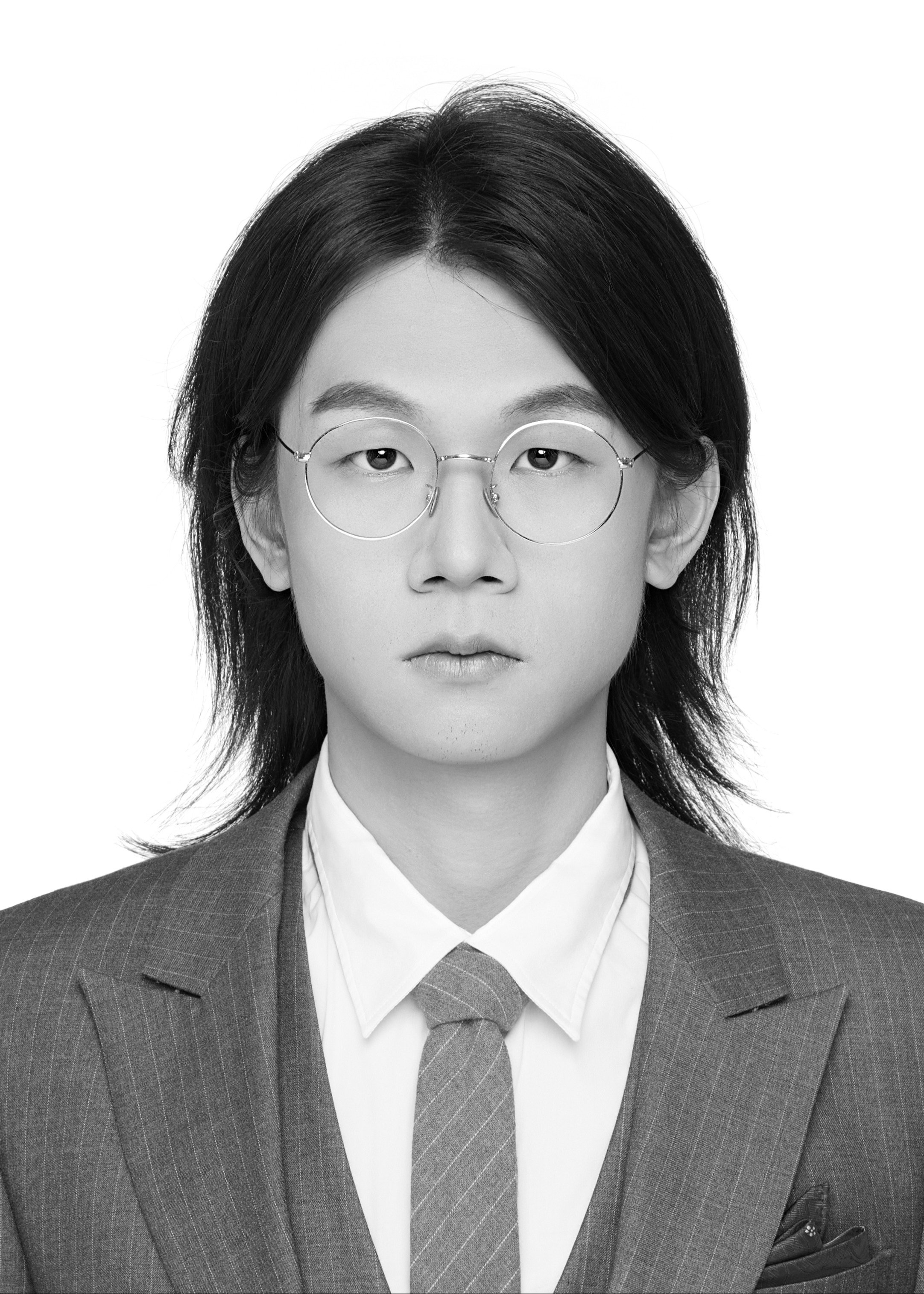}}]{Qi Zhang} received the B.S. degree from North University of China, Taiyuan, China, in 2022. He received an M.S. degree with the School of Computing Science, University Of Glasgow, Scotland, U.K. He is pursuing a Ph.D. degree at the University of Amsterdam, Netherlands. His current research interests include algorithms and systems for semantic SLAM in dynamic environments.
\vspace{-5mm}
\end{IEEEbiography}

\begin{IEEEbiography}
[{\includegraphics[width=1in,height=1.25in, clip]{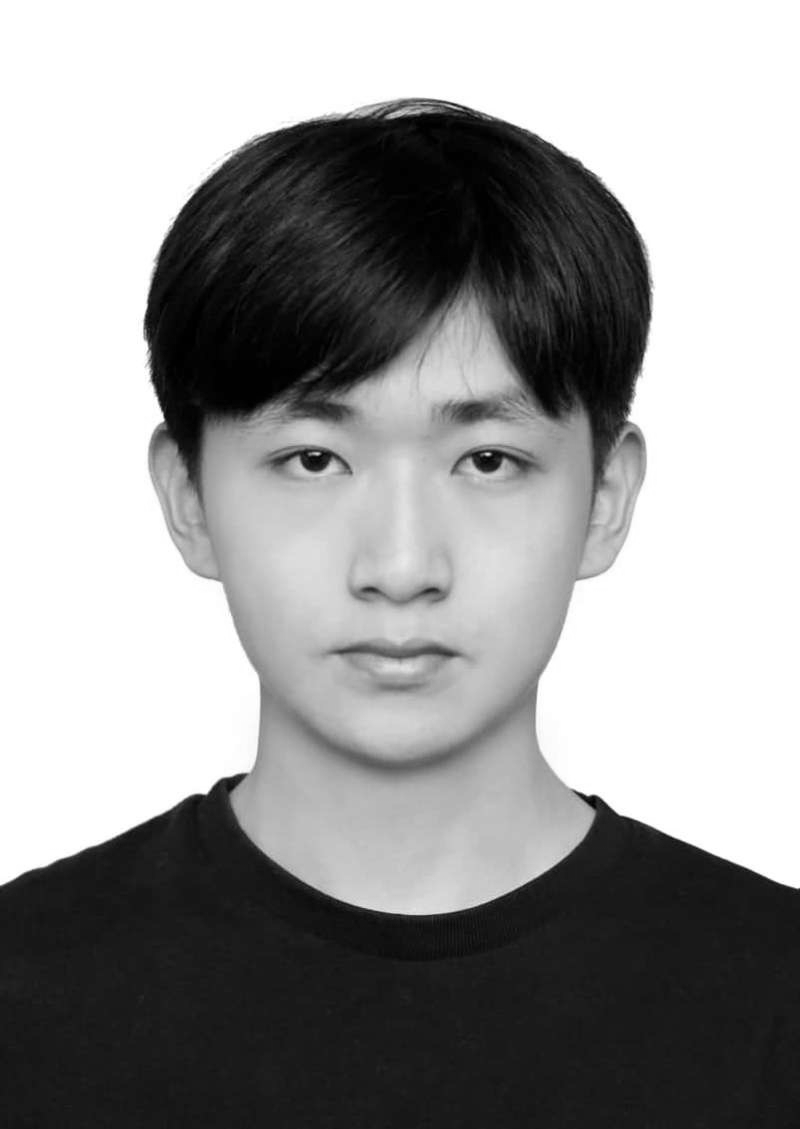}}]{Ziyang Ye}received his B.S. in Computer Science from the University of Adelaide, Adelaide, South Australia, Australia, in 2020. He is currently pursuing a M.S. degree in Artificial Intelligence and Machine Learning at the same institution. His research interests include 2D/3D computer vision tasks and reinforcement learning.
\vspace{-5 mm}
\end{IEEEbiography}

\begin{IEEEbiography}
[{\includegraphics[width=1in,height=1.25in, clip]{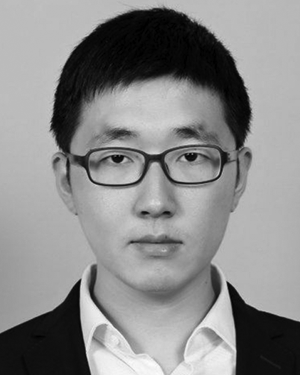}}]
{Hanyang Zhuang} (Member, IEEE) received the Ph.D. degree from Shanghai Jiao Tong University, Shanghai, China, in 2018. He was a Postdoctoral Researcher with Shanghai Jiao Tong University from 2020 to 2022. He is currently an Assistant Research Professor with Shanghai Jiao Tong University implementing research works related to intelligent vehicles. His research interest include AD/ADAS system design, high-precision localization, environment perception, and cooperative driving.
\vspace{-5 mm}
\end{IEEEbiography}

\begin{IEEEbiography}[{\includegraphics[width=1in,height=1.25in,clip,keepaspectratio]{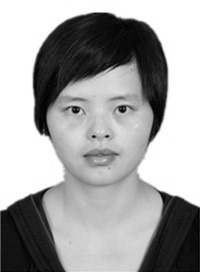}}]%
{Xianxian Zhao} received the Ph.D. degree in Electrical Engineering from the University of Birmingham in 2018. She is currently the Principle Investigator of the SEAI-RD\&D Programme at University College Dublin. Her research focuses on modelling and control of electric machines and converters, stability and power quality of power-electronic-based power systems, and model order reduction of large power systems.
\end{IEEEbiography}
\vfill



\end{document}